\pdfoutput=1
\documentclass[11pt]{article}
\usepackage[utf8]{inputenc}

\usepackage[preprint]{acl}

\usepackage{times}
\usepackage{latexsym}
\usepackage{textcomp}
\usepackage{upquote}           % straight quotes in code listings
\usepackage[T1]{fontenc}
\usepackage{xcolor}
\usepackage{listings}

\lstdefinestyle{plaintext}{
  basicstyle=\ttfamily\small,                         % equivalent to fontsize=\small
  breaklines=true,                                    % wrap long lines; can add visual cue below
  postbreak=\mbox{},% indicate broken line continuation. :contentReference[oaicite:0]{index=0}
  columns=fullflexible,                               % better line breaking / avoid overfull hboxes. :contentReference[oaicite:1]{index=1}
  keepspaces=true,                                    % preserve indentation/spaces. :contentReference[oaicite:2]{index=2}
  showstringspaces=false,
  frame=single,
  backgroundcolor=\color{gray!5},
  upquote=true,                                       % straight quotes; avoids curly quote issues. :contentReference[oaicite:3]{index=3}
}

\usepackage[utf8]{inputenc}

\usepackage{microtype}

\usepackage{inconsolata}

\usepackage{graphicx}
\usepackage{enumitem}
\usepackage{subcaption}  
\usepackage{amsfonts}
\usepackage{amsmath}
\usepackage{svg}
\usepackage{booktabs}
\usepackage{float}
\title{More Than a Score: Probing the Impact of Prompt Specificity on LLM Code Generation}

\author{Yangtian Zi \\
  Northeastern University \\
  \texttt{zi.ya@northeastern.edu} 
  \And
  Harshitha Menon \\
  Lawrence Livermore \\
  National Laboratory \\
  \texttt{harshitha@llnl.gov} 
  \And 
  Arjun Guha \\
  Northeastern University  \\
  \texttt{a.guha@northeastern.edu} 
  }

\newcommand{\partialordereval}{\textsc{PartialOrderEval}}

\begin{document}
\maketitle
\begin{abstract}

State‐of‐the‐art Large Language Models (LLMs) achieve high pass@1 on general benchmarks like HumanEval ~\citep{chen_EvaluatingLargeLanguage_2021} but underperform on specialized suites such as ParEval~\citep{nichols_CanLargeLanguage_2024b}.
Is this due to LLMs missing domain knowledge or insufficient prompt detail is given?
To answer this, we introduce \partialordereval{}, which augments any code generation benchmark with a partial order of prompts from minimal to maximally detailed.
Applying it to HumanEval and both serial and OpenMP subsets of ParEval, we measure how pass@1 scales with prompt specificity.
Our experiments with Llama‐3.x and Qwen2.5‐Coder demonstrate varying degrees of prompt sensitivity across different tasks, and a qualitative analysis highlights explicit I/O specifications, edge‐case handling, and stepwise breakdowns as the key drivers of prompt detail improvement.
\end{abstract}

\section{Introduction}

Since the emergence of Large Language Models (LLMs), there has been broad discourse about their effectiveness of code generation in both popular press and curated benchmarks.
Recent LLMs have become powerful code synthesis tools, achieving high pass@1 scores on common benchmarks such as MBPP~\citep{austin_ProgramSynthesisLarge_2021} and SWE-Bench~\citep{swebench}.
However, when evaluated on more niche programming domains--such as Bioinformatics~\citep{tang_BioCoder_2023}, Data Science~\citep{lai_DS1000NaturalReliable_2022} and parallel computing ~\citep{nichols_CanLargeLanguage_2024b}--models often fall short of expert-level performance, suggesting they are not yet a silver bullet for all programming challenges.

One interpretation is that current models simply lack the specialized knowledge to succeed in these more niche domains. 
Another possibility, however, is that they require more comprehensive contextual prompts than those provided by users (and, by extension, by existing benchmarks).
Indeed, beginning programmers frequently struggle to craft effective prompts due to an incomplete mental model of the information that needs to be conveyed and a limited grasp of model capabilities~\citep{nguyen_HowBeginningProgrammers_2024}. This raises a key question: could LLMs solve harder tasks—like parallel programming—if only they were guided with more detailed instructions?

\begin{figure*}[t]
    \centering
    \includegraphics[width=\textwidth]{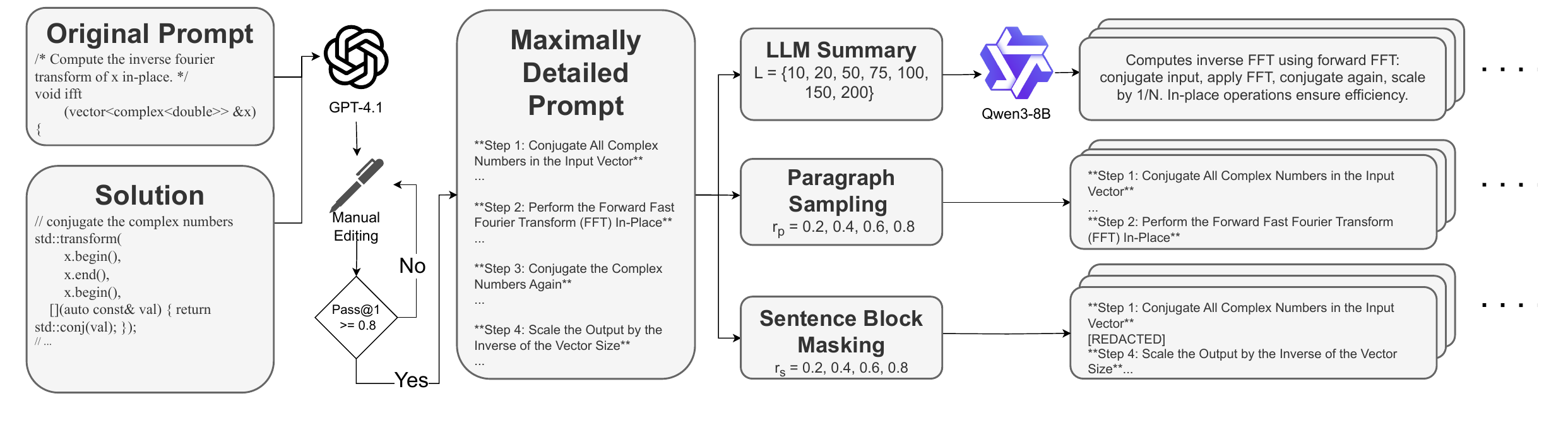}
    \caption{Overview of the \partialordereval{} prompt‐generation pipeline. Starting from an \emph{Original Prompt} and its reference \emph{Solution}, we use GPT-4.1 to draft a \emph{Maximally Detailed Prompt}, manually refine it to ensure a pass@1 $\geq$ 0.8, and designate it as $p_{top}$.
    From $p_{\mathrm{top}}$, we automatically derive three families of less‐detailed variants: (1) \emph{LLM Summarization} at word‐count limits $L\in\{10,20,50,75,100,150,200\}$, (2) \emph{Paragraph Sampling} at retention ratios $r_p$ and (3) \emph{Sentence Block Masking} with mask ratios $r_s$.
    These variants form the partial‐order prompt set used to evaluate model performance as a function of prompt specificity.}
    \label{fig:prompt-generation-procedure}
\end{figure*}

We introduce \partialordereval{}, a novel evaluation framework for LLM code generation that explicitly characterizes the spectrum of prompt specificity.
Rather then measuring performance under a single prompt, \partialordereval{} assesses each problem by generating multiple prompts of varying detail---ranging from minimal, high-level task descriptions to richly annotated, stepwise specifications (Figure \ref{fig:prompt-generation-procedure}).
Specifically, in \S \ref{sec:source-benchmarks} we detail our source benchmarks, in \S \ref{sec:problem-definition} we formalize the partial‐order framework, and in \ref{sec:dataset-construction} we describe how we construct the \partialordereval{} datasets.

This approach more faithfully captures how models perform under various realistic prompting strategies and reveals how sensitivity to prompt detail differs across models and task types.
By applying \partialordereval{} to both HumanEval~\citep{chen_EvaluatingLargeLanguage_2021} and the serial and OpenMP subsets of ParEval~\citep{nichols_CanLargeLanguage_2024b} and evaluating the resulting dataset against 2 series of models (\S \ref{sec:evaluation}), we show that this framework provides a more nuanced and accurate picture of model capabilities.
To give an example, we showed that LLM can indeed achieve higher pass@1 for two subsets of ParEval, even exceeding HumanEval figures---it suffices to add significantly more detail to the prompt.

Moreover, through analysis of our augmented prompts (\S \ref{sec:qualitative-analysis}), we identified key categories of information, such as detailed input/output specifications, explicit handling of edge cases, and structured implementation steps, that appear important for enabling LLMs to generate correct code. 
These findings suggest practical priorities for prompt engineering, helping developers focus on the details that yield the greatest return on effort.

We believe that \partialordereval{} will facilitate the development of more reliable prompting techniques, help developers identify the point of diminishing returns in prompt refinement, and ultimately drive progress toward LLMs that can truly assist programmers with minimal manual intervention.

\section{Related Work}

\paragraph{Multi-Prompt Evaluation in Code Generation}

Prior work has extended established benchmarks such as HumanEval~\citep{chen_EvaluatingLargeLanguage_2021} and MBPP~\citep{austin_ProgramSynthesisLarge_2021} by adding support for more programming languages~\citep{cassano_MultiPLEScalableExtensible_2022,athiwaratkun_MultilingualEvaluationCode_2022,orlanski_MeasuringImpactProgramming_2023,zheng_CodeGeeXPreTrainedModel_2023} and translating prompts to new natural languages~\citep{wang-etal-2023-mconala, peng-etal-2024-humaneval, raihan-etal-2025-mhumaneval}. Our work probes the \emph{depth} of prompting by systematically varying the level of detail in a prompt to quantify how prompt specificity affects LLM performance.

New programmers are worse at prompting than experts on coding tasks~\citep{nguyen_HowBeginningProgrammers_2024, mordechai-etal-2024-novicode, feldman_NonExpertProgrammersGenerative_2024, prather_WideningGapBenefits_2024,kazemitabaar_HowNovicesUse_2023a}. Close inspection of the StudentEval dataset~\citep{babe-etal-2024-studenteval} shows that new programmers often miss vital details in their prompts~\citep{lucchetti-etal-2025-substance}.

\paragraph{Prompting Strategies}
Several lines of work have explored prompting strategies to improve quality of LLM generation. 
Self-Refine~\citep{madaan2023selfrefine}, has the model critique and rewrite its own code over multiple turns.
Reflexion~\citep{shinn_ReflexionLanguageAgents} is a technique to improve quality of generated text through feedback in natural languages. 
\citet{gaoPALProgramaidedLanguage2023} improve problem solving performance, using prompts guiding LLMs to delegate problem-solving to generated Python programs. 
In contrast to iterative feedback and program‐aided techniques,\partialordereval{} employs a fixed hierarchy of prompt refinements, controlling the details, to precisely isolate how additional detail impacts code‐generation performance.

\paragraph{Multi-prompt for Evaluation Robustness}

Evaluating LLMs on a problem with multiple prompts have been advocated well before our work for the sake of robustness. \citet{mizrahi-etal-2024-state} found evidence that different prompt paraphrases leads to large performance disparity with various tasks including describing code, calling for aggregate metrics over diverse prompts.
PromptSet~\citep{kaiser2024promptset} mines over 61,000 real‐world developer prompts from open-source Python code, revealing broad variability in prompt effectiveness and suggesting that benchmarks should cover many prompt styles.
\citet{zhu_PromptRobustEvaluatingRobustness_2024} and~\citet{gu-etal-2023-robustness} investigated performance impacts of models when the prompt is perturbed.
Our work also employs different prompts to evaluate a given problem, but we systematically tune the level of detail in the prompts, capturing a new dimension in LLM evaluation.

\section{Building \partialordereval{} Benchmarks}

\partialordereval{} builds on any existing code-generations benchmark by systematically augmenting each problem with a suite of prompts that span a graduated spectrum of detail.
More precisely, we impose a \emph{partial order} over prompts based on a well defined detail measure.
During evaluation, an LLM is presented with each prompt in the set and its performance is recorded as a function of prompt specificity.

\subsection{Source Benchmarks}
\label{sec:source-benchmarks}

We start with two code synthesis benchmarks.
First, we use HumanEval~\citep{chen_EvaluatingLargeLanguage_2021}, which has been widely used for LLM evaluation. It is a benchmark of 164 Python problems. The largest and most capable LLMs achieve roughly 0.90 pass@1. 

The second benchmark that we use is ParEval~\citep{nichols_CanLargeLanguage_2024b}, which is a family of seven benchmarks that test model's ability to write scientific code using a variety of parallelism paradigms, including serial C++ (no parallelism), CUDA, AMD HIP, Kokkos~\citep{kokkos_citation}, and others.\footnote{ParEval also includes a translation task, e.g., translate serial C++ to use CUDA, which we do not use in this paper.} In this paper, we use the serial and OpenMP (OMP) problems.
OpenMP is an API for shared-memory parallel processing code.
Each subset has 60 problems (120 total) that exercise the ability of models to write code to solve scientific and parallel‐computing tasks, e.g., fast fourier transforms, prefix sums, and graph operations.

Whereas HumanEval is saturated, both ParEval-Serial and ParEval-OpenMP are significantly harder. For example, while Qwen2.5-Coder-14B-Instruct~\citep{hui_qwen2_2024} gets 0.866 on HumanEval, it gets 0.800 and 0.667 on ParEval-Serial and ParEval-OpenMP respectively (Table \ref{tab:model_performance}).

But, \emph{what really makes ParEval harder?} Could it be that with just a little more detail in the prompts, success rates on ParEval would improve dramatically? We formalize this problem and study it in depth below.

\subsection{Problem Definition}
\label{sec:problem-definition}

Consider a fixed model $\mathcal{M}$ under evaluation. A coding benchmark consists of pairs $(p, \mathsf{eval}_p)$, where $p$ is a prompt and $\mathsf{eval}_p$ denotes the associated hidden tests. 
We first construct a \textit{maximally detailed prompt} $p_\mathit{top}$ from $p$, constructed such that the model $\mathcal{M}$ generates correct solutions with high pass@1. Formally, we require $\mathbb{E}(\mathsf{eval}_p(\mathcal{M}(p_\mathit{top}))) \ge \tau$.
In this paper we use $\tau = 0.8$ as the threshold.

We define a prompt detail metric $D$ that measures the level of detail within prompts.
Given this metric, we identify the top prompt $p_\mathit{top}$ with maximum details and the bottom prompt $p_\mathit{bot}$ with minimal details.
We then construct an intermediate set of prompts $P$, such that for all $p \in P$, $p_\mathit{bot} <_D p <_D p_\mathit{top}$ when ordered by $D$.
Thus $D$ imposes a partially ordered set of prompts $P^* = P\cup \{p_\mathit{top}, p_\mathit{bot}\}$, with $p_\mathit{top}$ as the maximum and $p_\mathit{bot}$ as the minimum as defined by $D$.
The question that we ask is the following: given any two prompts $p_1, p_2 \in P^*$ satisfying the order $p_1 <_D p_2$, does increasing prompt detail always yield non-decreasing model correctness, that is,  
$\mathbb{E}(\mathsf{eval}_p(\mathcal{M}(p_1))) \le \mathbb{E}(\mathsf{eval}_p(\mathcal{M}(p_2)))$?

As we show in Section~\ref{sec:evaluation}, as long as the metric $D$ is reasonably defined, this empirically generally holds for three programming benchmarks and several models.

\subsection{Dataset Construction}
\label{sec:dataset-construction}

\begin{figure*}[ht]
    \centering
    \includegraphics[width=\textwidth]{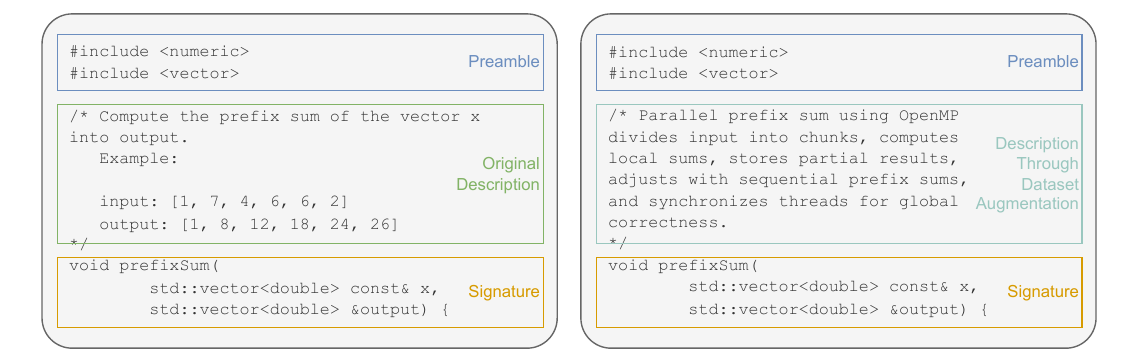}
    \caption{The three components of a prompt: \emph{Preamble}, \emph{Description} and \emph{Signature}, illustrated by comparing the original ParEval prompt for problem \texttt{30\_scan\_prefix\_sum} (left) with its PO-ParEval-OMP variant produced by a 25‐word LLM summary (right).
    In \partialordereval{}, only the \emph{Description} section is replaced when generating prompt variants.}
    \label{fig:original-vs-new}
\end{figure*}

Each HumanEval and ParEval prompt comprises of three components:
\begin{itemize}
  \item \textbf{Preamble}: any necessary imports or helper functions.
  \item \textbf{Description}: the natural‐language problem statement.
  \item \textbf{Signature}: the function signature (including argument types for ParEval).
\end{itemize}

During evaluation, the LLM is tasked with generating the function body immediately following the header.  Our augmentation strategies operate on the \textbf{description} component, producing alternative prompt texts that replace the original problem statement.  See Figure~\ref{fig:original-vs-new} for an illustration of this procedure.

We construct \partialordereval{} variants from HumanEval and Pareval: \textsc{PO-HumanEval} (164 problems), \textsc{PO-ParEval-Serial} (60 problems) and \textsc{PO-ParEval-OMP} (60 problems).
See an illustration of the entire dataset generation procedure in Figure \ref{fig:prompt-generation-procedure}.

To start, for each problem in HumanEval, ParEval-Serial and ParEval-OMP, we generated the maximally detailed prompt $p_\mathit{top}$ using the original dataset's prompt and solution using GPT-4.1-2025-04-14~\citep{openai_gpt41_2025} (See Appendix~\ref{sec:app-prompt-max-detail} for the prompt used).
We made sure that generated $p_\mathit{top}$ does not have the complete solution verbatim in the comment.
We verify that the prompts can achieve $\text{pass@1}\geq0.8$ for all problems with Qwen2.5-Coder-14B~\citep{hui_qwen2_2024}.
We found that 6 ParEval-Serial and 19 ParEval-OMP prompts needs to be manually edited or re-generated to make the pass@1 threshold.
Then, from these prompts, we derive less-detailed prompts via three \textit{augmentation strategies}:

\begin{enumerate}
\item \textbf{LLM Summarization.} We prompt Qwen3-8B \cite{qwen3technicalreport} to condense generated maximally detailed prompts into summaries of up to $L=10, 25, 50, 75, 100, 150$ and $200$ words.
The metric $D$ here is defined by the word limit of the prompts,
and summaries with less word limit naturally contains less details.

\item \textbf{Paragraph Sampling.} We randomly sample paragraphs from the seed description at ratios $r_p = 0.2, 0.4, 0.6, 0.8$. 
That is, we randomly choose a subset of paragraphs of the given $r_p$, and more information is included as $r_p$ is increased.

\item \textbf{Sentence Block Masking.}  For each mask ratio \(r_s \in \{0.2, 0.4, 0.6, 0.8\}\), we remove a contiguous block of \(r_s\!\times\!100\%\) of sentences. 
This block is placed at four evenly spaced start positions---at the beginning, two midpoints, and the end---yielding four variants that progressively strip more detail as \(r_s\) increases.
\end{enumerate}

For LLM summarization, we generate only a single prompt per word‐limit, since summaries at the same length tend to be paraphrases of each other.
In contrast, paragraph sampling and sentence block masking introduce randomness in which content is retained or removed. To account for this variability, we produce 4 distinct prompts at each sampling ratio or masking level, ensuring a consistent degree of detail while capturing differences in content location.

We place a full set of augmented prompts for a problem in Appendix \ref{sec:app-sample-prompts}.

Together, these techniques enable \partialordereval{} to probe model performance \emph{across} and \emph{within} detail levels, revealing fine‐grained insights into LLMs' sensitivity to prompt engineering.
Thus each problem yields $41$ distinct prompts, consisting of:
\begin{itemize}
    \item The minimally detailed prompt, containing only the original function signature and any required preamble with no description, $p_\mathit{bot}$.
    \item The maximally detailed prompt $p_\mathit{top}$.
    \item $39$ Intermediate prompts obtained via our three augmentation strategies. LLM summarization have $1$ intermediate prompt per $L$, yielding $7$ prompts. Paragraph sampling and sentence block masking have $4$ intermediate prompts per $r_p$ and $r_s$, yielding $32$ distinct prompts.
\end{itemize}

\begin{table*}[ht]
\centering
\begin{tabular}{llccc}
\toprule
Model & Size & \multicolumn{3}{c}{Pass@1 ($n=1$, Greedy Sampling)} \\
\cmidrule(lr){3-5}
      &      & HumanEval     & ParEval-Serial  & ParEval-OMP    \\
\midrule
\multicolumn{5}{c}{\textbf{Qwen 2.5 Coder Series}} \\
Qwen2.5-Coder-1.5B-Instruct  & 1.5B & 0.659 & 0.517 & 0.300 \\
Qwen2.5-Coder-3B-Instruct    & 3B   & 0.762 & 0.717 & 0.433 \\
Qwen2.5-Coder-7B-Instruct    & 7B   & 0.774 & 0.817 & 0.517 \\
Qwen2.5-Coder-14B-Instruct   & 14B  & 0.866 & 0.800 & 0.667 \\
\midrule
\multicolumn{5}{c}{\textbf{Llama 3.x Series}} \\
Llama-3.2-1B-Instruct  & 1B   & 0.305 & 0.283 & 0.100 \\
Llama-3.2-3B-Instruct  & 3B   & 0.506 & 0.467 & 0.183 \\
Llama-3.1-8B-Instruct  & 8B   & 0.622 & 0.583 & 0.367 \\
Llama-3.3-70B-Instruct & 70B  & 0.744 & 0.750 & 0.533 \\
\bottomrule
\end{tabular}
\caption{Instruction‐tuned LLMs evaluated in our study, listed by family and parameter count, with their corresponding pass@1 scores on the original HumanEval, ParEval‐Serial and ParEval-OMP  source datasets.}
\label{tab:model_performance}
\end{table*}

\section{Evaluation}
\label{sec:evaluation}

We measure pass@1---reported as a decimal between 0 and 1---for each prompt variant by running the generated code on the hidden test suites and averaging over all problems at that specificity level.
We plot these averages against prompt detail to produce performance curves that show how accuracy changes as prompts become more detailed.

\subsection{Models and Parameters}
\label{sec:models-and-params}

We benchmark two families of instruction-tuned LLMs: Qwen 2.5 Coder~\citep{hui_qwen2_2024} and Llama 3.x~\citep{grattafiori2024llama3herdmodels}. See Table~\ref{tab:model_performance} for a list of models we used, their size and Pass@1 score for the source datasets for reference.
By selecting two series of open-weights models with model size spanning from 1B to 70B parameters.
This selection allow us to isolate how model scale and architectural differences influence performance.

\subsection{Metrics}
\label{sec:metrics}

We report pass@1 values as decimal numbers from 0 to 1, instead of percentages.
To visualize how sensitive the models are to varying degrees of prompt detail, we plot the pass@1 values across different prompt specificity, creating a \emph{performance curve} for each model-benchmark pair.
Such curves enable us to identify the minimum prompt‐detail level required for models to reliably produce correct solutions and quantify the performance improvement as prompts become increasingly informative.

\begin{figure*}[t]
    \centering
    \includegraphics[width=\textwidth]{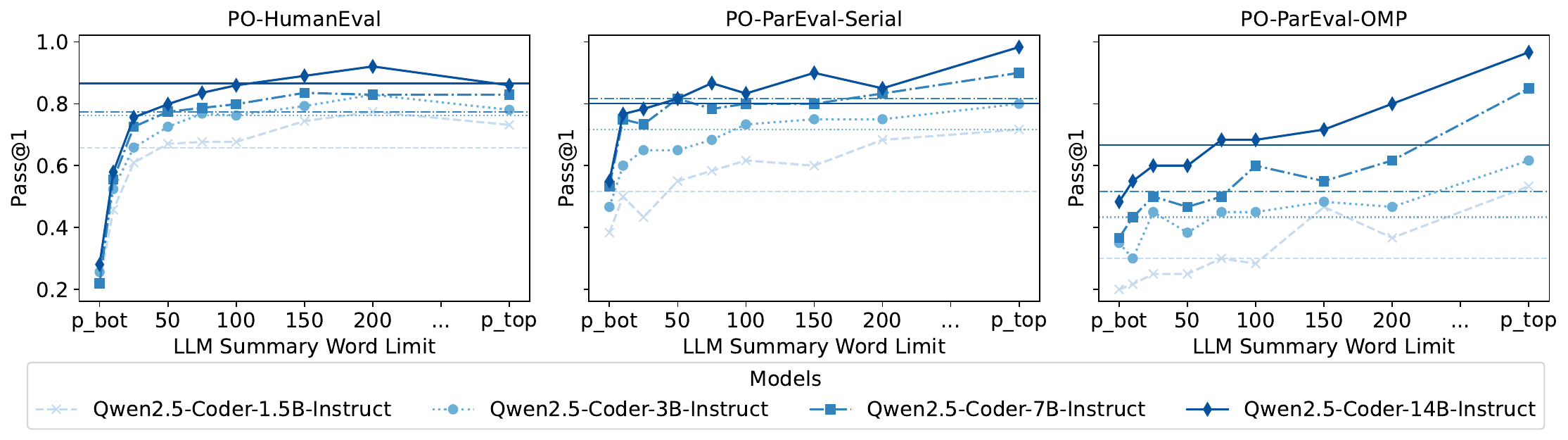}
    \includegraphics[width=\textwidth]{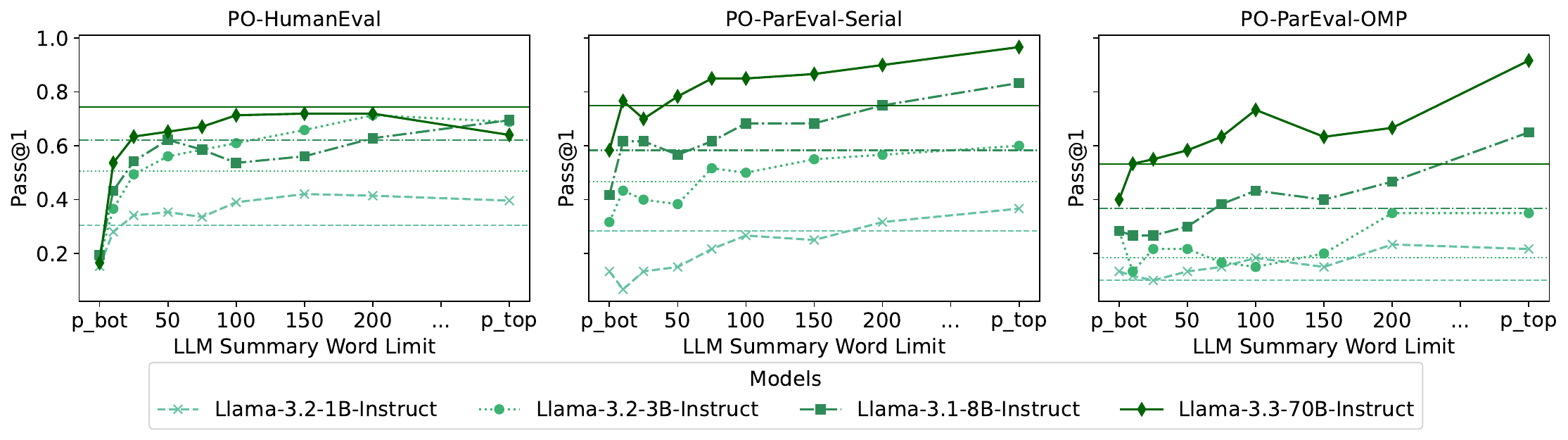}
    \caption{Pass@1 versus prompt detail (word limit) for LLM summarization across all \partialordereval{} datasets and model sizes. 
    The y‐axis shows average pass@1, and the x‐axis shows word limits from the minimal prompt ($p_{bot}$) through various summary lengths to the fully detailed seed ($p_{top}$).
    Panels (left to right) display results on \textsc{PO-HumanEval}, \textsc{PO-ParEval-Serial}, and \textsc{PO-ParEval-OMP}.
    Each line style corresponds to a different Qwen or Llama model size, and horizontal lines denote each model’s performance on the original benchmark prompt.
    Note how curves plateau around $L=100$ for \textsc{PO-HumanEval} but continue rising for the ParEval variants.}
    \label{fig:qwen-llm-summary}
\end{figure*}

\subsection{Results}
\label{sec:results}

To present the results, we will show a representative subset of performance curves.
Each curve shows a model’s average pass@1 (vertical axis) as prompt specificity increases (horizontal axis), starting from a minimal description on the left and ending with the fully detailed seed prompt on the right.
Line styles and markers distinguish different models or augmentation strategies.
A rising segment indicates that more information helps the model, a plateau shows that further detail no longer improves accuracy, and any dips suggest potential information overload or confusion.
We illustrate our findings using Qwen results primarily but observed similar patterns in the Llama models.
We place complete results of our experiments in Figure~\ref{fig:main-results} in the Appendix.

\paragraph{LLM Summary}

Figure~\ref{fig:qwen-llm-summary} presents pass@1 as a function of summary length $L$ for \textsc{PO-HumanEval}, \textsc{PO-ParEval-Serial}, and \textsc{PO-ParEval-OMP}.  
On \textsc{PO-HumanEval}, all models show rapid gains from $p_\mathit{bot}$ to $L=50$, tapering off toward $L=100$ where performance nearly equals that at $p_\mathit{top}$, and exhibit a slight decline beyond $L=200$.
For instance, Qwen2.5-Coder-14B-Instruct climbs from 0.280 at $p_\mathit{bot}$ to 0.799 at $L=50$ and reaches 0.860 by $L=100$ (matching $p_\mathit{top}$).
Its performance plateaus for $L = 200$ at 0.921, but declines back to 0.860 at $p_{top}$. 

By contrast, \textsc{PO-ParEval-Serial} sees more gradual improvements—Qwen2.5-14B-Instruct reaches 0.867 at $L=75$ and 0.900 at $L=150$, yet remains below its $p_\mathit{top}$ score of 0.983.  \textsc{PO-ParEval-OMP} is the most challenging: the same model rises slowly from 0.483 at $p_\mathit{bot}$ to 0.800 at $L=200$, but never attains its $p_\mathit{top}$ accuracy of 0.967.

All of the above thrends can also be re-confirmed with Llama models.

Across all benchmarks, larger models in parameters consistently outperform smaller ones at every specificity level.  The contrast between swift convergence on \textsc{PO-HumanEval} and protracted gains on ParEval variants underscores how prompt sensitivity can signal task difficulty.  Additionally, the slight performance drop beyond $L=200$ on \textsc{PO-HumanEval} suggests that overly verbose prompts may introduce redundancy or cognitive overload, marginally impairing code‐generation accuracy.

\paragraph{Paragraph Sampling and Sentence Block Masking}

\begin{figure*}[t]
    \centering
    \includegraphics[width=\textwidth]{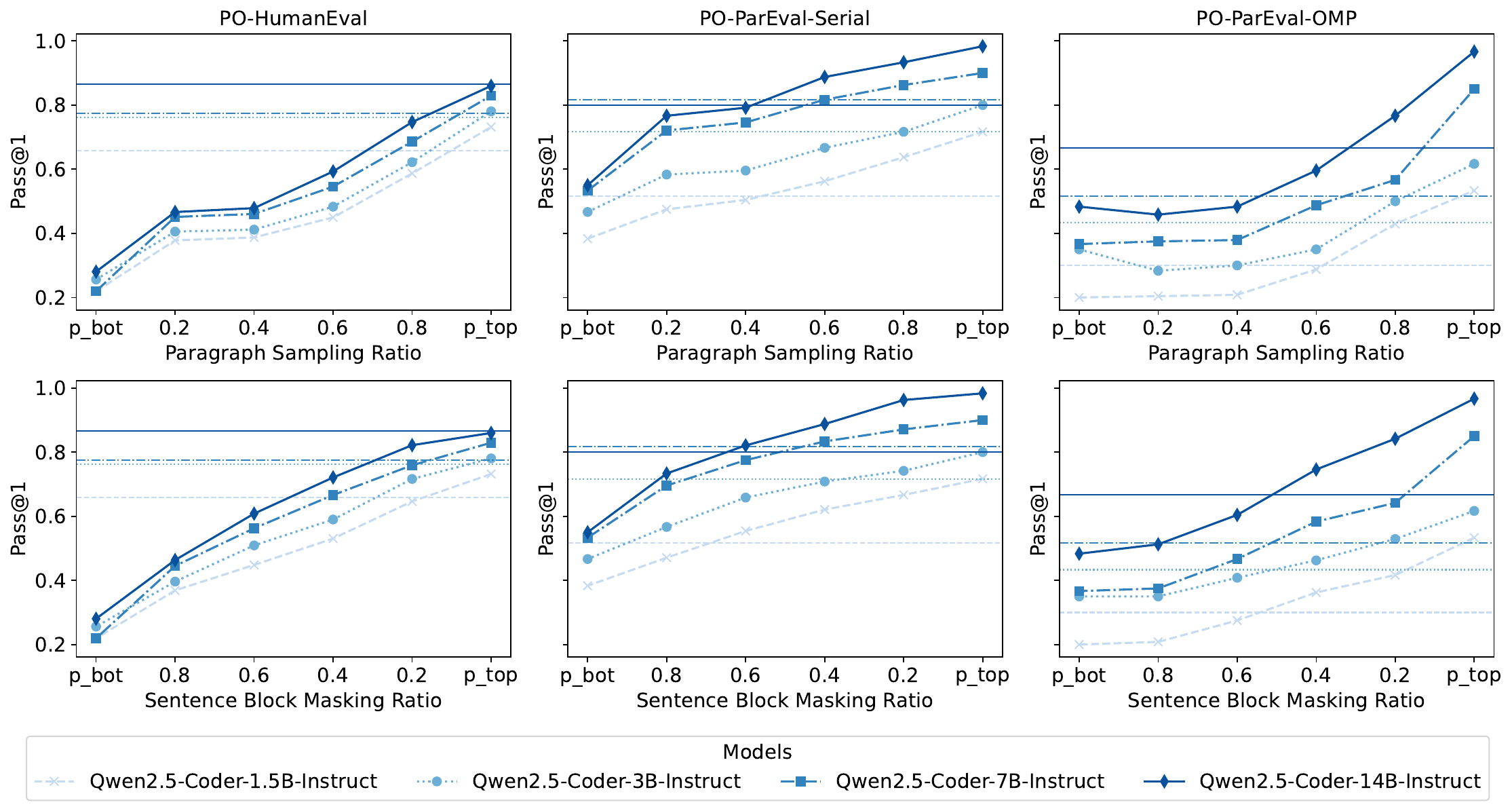}
    \caption{Pass@1 versus prompt detail for Paragraph Sampling and Sentence Block Masking augmentation across all \partialordereval{} datasets and Qwen models.
    As in Figure~\ref{fig:qwen-llm-summary}, the y-axis shows pass@1, x-axis ranges from minimal to maximal prompt detail (note that for Sentence Block Masking, detail increases as $r_s$ decreases), and horizontal lines stand for original dataset performance.
    Each line style represents a different model size, illustrating that larger models achieve higher accuracy with less prompt specificity. Similar trends hold for the Llama series of models (see Appendix Figure \ref{fig:main-results}.)}
    \label{fig:qwen-p-and-s}
\end{figure*}

Figure \ref{fig:qwen-p-and-s} shows that both paragraph sampling and sentence‐block masking yield very similar trends: for all three datasets, pass@1 steadily improves as prompt detail increases (i.e., higher $r_p$ or lower $r_s$), and larger models consistently outperform smaller ones at every level.
However, the gap between big and small models remains modest on \textsc{PO-HumanEval} but widens substantially on the ParEval variants.
For instance, at $r_p=0.8$ in paragraph sampling, Qwen2.5-Coder-14B-Instruct achieves 0.747 on \textsc{PO-HumanEval}—just 0.16 above the 1.5B model’s 0.587—but the gap grows to 0.296 on \textsc{PO-ParEval-Serial} (0.933 vs.\ 0.637) and 0.338 on \textsc{PO-ParEval-OMP} (0.767 vs.\ 0.429). 

These results confirm that more detailed prompts not only boost overall accuracy but also serve as a sensitive probe of model capability: stronger models require less prompt specificity to reach a given performance level, especially on more challenging, domain‐specific tasks.

\paragraph{Comparing to Original Prompts}

Figures \ref{fig:qwen-llm-summary} and \ref{fig:qwen-p-and-s} also show that, at the maximally detailed prompt $p_{top}$, models substantially outperform their scores on the original ParEval prompts.
For example, Qwen2.5‐Coder-14B reaches 0.983 on \textsc{PO-ParEval-Serial} and 0.967 on \textsc{PO-ParEval-OMP}—versus only 0.800 and 0.667 on the unaugmented ParEval benchmarks.
By contrast, performance on HumanEval remains essentially unchanged ($\approx0.86$) whether using the original prompt or $p_{top}$. 

This difference indicates that a increase in contextual detail can unlock dramatic improvements---up to a 0.30 absolute gain on ParEval-OMP---whereas extra detail yields diminishing returns on the relatively easier HumanEval tasks.
Notably, at intermediate specificity levels (e.g.\ 50–100 words in the LLM summarization), \textsc{PO-ParEval} performance already surpasses the original prompt, suggesting that only a moderate amount of additional instruction is required to outperform the original benchmarks.
However, achieving near‐perfect pass@1 still requires substantial additional prompt engineering effort.

\paragraph{Summary: ParEval is More Challenging for LLMs}

Our evaluation shows that models converge to their maximal‐detail performance much more slowly on ParEval than on HumanEval: on both ParEval‐Serial and ParEval‐OMP, pass@1 increases gradually and never quite reaches the $p_{top}$ ceiling that HumanEval models hit by around 100 words of detail.
Moreover, whereas the gap between large and small models on HumanEval is relatively small, it widens dramatically on ParEval, making these datasets stronger discriminators of model capability.
Finally, significant performance improvements achieved through enhanced prompt specificity---up to 0.30 absolute gains for ParEval-OMP---underscore that ParEval tasks, especially in parallel programming, are inherently more challenging and sensitive to prompt specificity than HumanEval.

\section{What Prompt Details Matter?}
\label{sec:qualitative-analysis}

\begin{figure}[h]
    \centering
    \includegraphics[width=\linewidth]{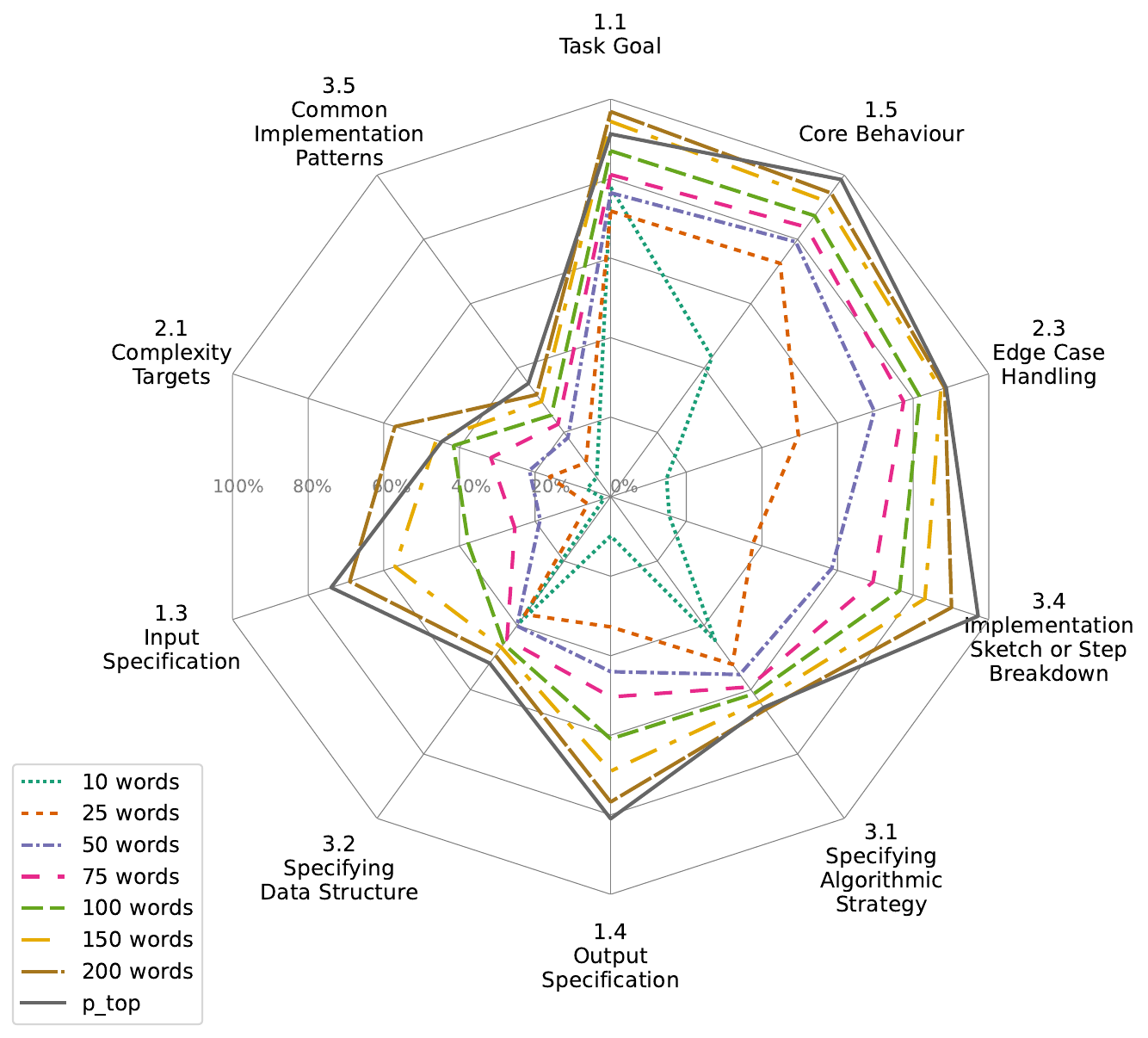}
    \caption{Radar chart of the top 10 taxonomy themes as they appear in LLM‐summarized prompts of varying length, including the fully detailed prompt ($p_{top}$).
    Each axis corresponds to one theme, and concentric gridlines denote percentage increments (e.g., $20\%$, $40\%$, $60\%$) of prompts containing that theme at each summary length.  The radial position of each marker indicates the proportion of prompts (out of 284 per length) in which the theme occurs, averaged across all problems.}
    \label{fig:spider-top-10}
\end{figure}

Beyond quantitative pass@1 measurements, we performed a qualitative analysis focusing specifically on prompts generatd through the \emph{LLM Summarization} augmentation strategy.
Our goal is to better understand  \emph{which} categories of prompt details most strongly contribute to the observed performance gains as prompt specificity increases.

To systematically analyze prompt contents, we first developed a structured taxonomy with guidance from o3-2025-04-16~\citep{OpenAI2025o3_o4mini} .
This taxonomy organizes prompt details into four high-level \emph{categories}: \textit{1. Functional Specification, 2. Constraints and Robustness, 3. Solution Structure and Design Guidance, 4. Verification and Integration}.
Each category comprises multiple detailed \emph{themes}, numbered as under these themes. for example, the \textit{Functional Specification} category includes themes such as \textit{1.1 Task Goal}, \textit{1.3 Input Specification}, and \textit{1.5 Core Behavior}.
Themes are not mutually exclusive, and a single prompt may be annotated with several overlapping themes. 
A complete version of the taxonomy, along with concise descriptions of each theme, is provided in Appendix~\ref{sec:taxonomy-of-prompt-details}.

Using claude-sonnet-4-20250514~\citep{anthropic2025systemcard}, we annotated all LLM Summary prompts at different word limits according to our taxonomy.
Specifically, each prompt was labeled with all applicable instruction themes identified within its text. 
We then select the top 10 themes ranked by the number of occurrences averaged by length and plot it in Figure~\ref{fig:spider-top-10}.
We place our prompting strategy for LLM-assisted prompt annotation in Appendix \ref{sec:prompting-qual}.

We found that \textit{1.1 Task Goal}, which signifies the required outcome of the program, consistently appeared in nearly all prompts regardless of their length, highlighting it as a foundational component of effective prompting.
More intriguingly, certain themes exhibited substantial growth in frequency as prompt specificity increased.
Specifically, the themes \textit{1.5 Core Behavior}, \textit{1.3 Input Specification}, \textit{1.4 Output Specification}, \textit{2.3 Edge Case Handling}, and \textit{3.4 Implementation Sketch or Step Breakdown} significantly rose in prominence at higher word limits.
The increased presence of these themes suggests they might play particularly influential roles in enabling LLMs to achieve higher correctness, possibly by offering more structured and explicit guidance on how the task should be approached, clarifying requirements, and reducing ambiguity.
Other themes among the top 10 seemed to exhibit less growth: \textit{3.1 Specifying Algorithmic Strategy}, \textit{3.2 Specifying Data Structure} and \textit{3.5 Common Implementation Patterns}.
This phenomenon suggests that these themes might be implicitly understood by LLMs--for example, by describing an implementation sketch, all three more modest gained themes might already be implicitly mentioned.
Mentioning them again might provide only marginal improvement to correctness.

Our preliminary observations indicate potential avenues for effective prompt design.
In particular, they suggest prioritizing explicit instruction on expected input/output formats, critical problem-solving steps, handling of edge cases, and providing structured breakdowns or pseudo-code as prompts are elaborated.
Future work could involve targeted ablation studies to quantify the impact of each identified theme individually, thereby confirming and further refining these recommendations.

\section{Conclusion}

In summary, we introduced \partialordereval{}, a framework  for evaluating models across prompts ranging from minimal to fully detailed. 
Through experiments on HumanEval, ParEval-Serial and ParEval-OMP, that increased prompt specificity consistently improves pass@1, though improvements vary by task complexity.
Furthermore,  ParEval tasks, especially parallel variants, converge more slowly and show greater sensitivity to prompt detail, making them more effective benchmarks than HumanEval for distinguishing model capabilities.
We demonstrated that \partialordereval{} provides a more nuanced differentiation of model capabilities than single-prompt pass@1 number alone.
Additionally, our qualitative analysis identifies key instruction elements that significantly improve correctness, offering concrete guidance for prompt engineering.

We envision \partialordereval{} as a first step toward a more holistic suite of evaluation tools for LLM-driven programming, one that accounts for both the breadth of tasks and the depth of prompt design. Future work can extend this framework to interactive prompting, dynamic feedback loops, and domain‐specific benchmarks, ultimately advancing our understanding of how to leverage LLMs as effective coding assistants.

\section*{Limitations}

While \partialordereval{} offers a more fine‐grained view of LLM code‐generation capabilities, there are serval limitations:

\paragraph{Alternative Augmentation Strategies}
We explore three augmentation techniques: LLM‐based summarization, paragraph sampling, and sentence‐block masking.
However, these represent only a small portion of the possible ways to vary prompt detail. Alternative strategies might interact differently with model architectures or training regimes, and could yield distinct model performance versus sensitivity profiles.

\paragraph{Qualitative Analysis}
Our taxonomy‐based analysis (\S \ref{sec:qualitative-analysis}) of prompt content uses an LLM to generate the taxonomy and classify text. The taxonomy is potentially biased by choice of model, but we manually verified a sample of results to ensure validity.

\paragraph{Programming Languages Used}
Our experiments employ Python (for HumanEval) and C++ (for ParEval), both of which are classified as high‑resource programming languages. Prior studies have shown that high‑resource programming languages tend to outperform low‑resource ones in code generation tasks~\citep{multipl_e,cassano_KnowledgeTransferHighResource_2024}.
Investigating how \partialordereval{} performance curves vary for low‑resource languages is future work.

\section*{Acknowledgments}
This material is based upon work supported by the U.S. Department of Energy, Office of Science, Office of Advanced Scientific Computing Research, through solicitation DE-FOA-0003264, "Advancements in Artificial Intelligence for Science," under Award Number DE-SC0025598 and contract DE-AC52-07NA27344. This work was performed under the auspices of the U.S. Department of Energy by Lawrence Livermore National Laboratory (LLNL) under Contract DE-AC52-07NA27344 (LLNL-CONF-2009663).

% Bibliography entries for the entire Anthology, followed by custom entries
\bibliography{anthology,custom}
% Custom bibliography entries only
%\bibliography{custom}

\appendix
\section{Experimental Details}

\subsection{Computational Information and Budget}

All experiments in Section \ref{sec:evaluation} were run in a server with Intel Xeon Gold 6342 CPU at 2.80GHz with 96 cores and 4 Nvidia H100 GPUs.
Each model and source dataset pair is evaluated on 18 data points (7 LLM summaries, 4 paragraph samplings, 4 sentence‐block maskings, 1 minimal prompt, 1 maximal prompt, and 1 original prompt).  With each run lasting 10 minutes ($\approx0.167$ GPU‐hours) and using 2 GPUs (4 GPUs for Llama-3.3-70B), the total compute is:

\begin{itemize}
  \item \textbf{Qwen-2.5-Coder series (4 models on 2 GPUs):}  
    \(4 \times 18 \times 0.167 \times 2 \times 3 \approx 72\) GPU-hours  
  \item \textbf{Llama-3.x series small (3 models on 2 GPUs):}  
    \(3 \times 18 \times 0.167 \times 2 \times 3 \approx 54\) GPU-hours  
  \item \textbf{Llama-3.3-70B (1 model on 4 GPUs):}  
    \(1 \times 18 \times 0.167 \times 4 \times 3 \approx 36\) GPU-hours  
\end{itemize}

In total, our evaluation consumes on the order of \(162\) GPU-hours.Additionally, we spend 10 GPU hours on experimenting and prototyping LLM Summarization using Qwen3-8B. Hence, for this work we spend at around $172$ GPU-hours.

Program correctness verification ran on CPU:  
\begin{itemize}
  \item HumanEval tests: $\approx8$ s per data point (24 threads)  
  \item ParEval-Serial tests: $\approx1.5$ min per data point  
  \item ParEval-OMP tests: 5--30 min per data point (due to potential deadlocks)  
\end{itemize}
Overall, end‐to‐end evaluation for each model–dataset pair required roughly 8–12 CPU hours under this setup.

\subsection{API Usage and Cost}
\label{sec:api-cost}

In addition to the open‐weight LLMs, we leverage GPT-4.1 and Claude Sonnet 4 via their respective APIs for summarization and taxonomy labeling.
By batching requests where possible, our total API expenditure remains under \$10 USD.
We also accessed o3-2025-04-16 through the ChatGPT interface under a \$20 USD/month subscription plan.  

\subsection{Model Inference Parameters}
\label{sec:inference-params}

For generating maximally detailed prompt (\S \ref{sec:dataset-construction}), we used GPT-4.1 with temperature = 0.7 and top\_p = 0.95.

All code‐generation evaluations in \S\ref{sec:evaluation} use greedy decoding (temperature = 0).  

For the LLM Summarization augmentation (\S\ref{sec:dataset-construction}), we invoke Qwen3-8B with temperature = 0.2 and top\_p = 0.95.  

In the qualitative analysis (\S\ref{sec:qualitative-analysis}), taxonomy labels are obtained via single‐shot prompts to ChatGPT (o3-2025-04-16) and Claude Sonnet 4, both using their default inference settings. 

\subsection{Prompting Procedure for Evaluation}

For HumanEval, we prepend each prompt with a fixed instruction template (see Appendix~\ref{sec:app-prompt-humaneval}). In ParEval, we follow the authors' original prompt design~\citep{nichols_CanLargeLanguage_2024b}, injecting our augmented description in place of their problem statement.
All prompts include any necessary import statements or helper functions to ensure that generated completions can be executed directly.
\section{Prompt for Generating Maximally Detailed Prompt}
\label{sec:app-prompt-max-detail}

\begin{lstlisting}[style=plaintext]
# System Prompt

You are an expert in <C++ or Python, depending on the language>. The user will provide a code snippet, and the code snippet will contain a function whose name starts with **correct**. You should produce a description of that specific function as a hint for how to implement it. You should only describe the function itself (not the surrounding code), and refer to it as "the function" or "the function provided," never by its actual name. If the implementation requires helper functions, your description should still be self-contained.

You should describe the function step by step, and include all the details of the function.

# User Prompt

<Code snippet>
\end{lstlisting}

\section{Prompt for HumanEval Evaluation}
\label{sec:app-prompt-humaneval}

\begin{lstlisting}[style=plaintext]
# System Prompt
You are a helpful programming expert that generates clean, idiomatic Python code
according to the user's instructions.

# User prompt

Complete the following Python function. Write only the body of the function, do
not include any other text, e.g. do not include any other text before or after
the function body. Enclose your solution in ```python and ```.

<HumanEval Prompt here>
\end{lstlisting}

\section{Prompting for Qualitative Analysis}
\label{sec:prompting-qual}

We designed two distinct prompt templates for our qualitative analysis:  
\begin{enumerate}
  \item \textbf{Taxonomy Generation Prompt:} Issued to o3-2025-04-16 via the ChatGPT interface, this prompt solicited a structured taxonomy based on sample LLM‐summarized prompts.  
  \item \textbf{Theme Labeling Prompt:} Sent to Claude Sonnet 4 through its API, this prompt presented each LLM‐summarized prompt instance and asked the model to assign all applicable taxonomy themes.
\end{enumerate}

\subsection{Taxonomy Generation Prompt}

\begin{lstlisting}[style=plaintext]
# Query

[all_llm_summary.txt, containing llm summaries at 7 lengths for all problems]

Here is a file containing a lot of solution descriptions for nearly 300 coding problems with different lengths. You need to look into each and every descriptions, and produce a generalized 2-level taxonomy system (with 4-6 top-level categories, but you can have as many sub-categories as you can)  that generalizes the possible information portrayed in the descriptions. 

# Reply from Model

To generate a well-structured 2-level taxonomy from the coding problem solution descriptions, I'd like to confirm a few things:

1. **Focus Dimensions**: Should the taxonomy categorize solutions based on the type of **algorithms**, **problem domains**, **data structures**, **implementation techniques**, or a combination of these?
2. **Desired Format**: Do you want the output in a visual hierarchy (like a tree or outline) or as a structured list?
3. **Intended Use**: Is the taxonomy meant for tagging/classification, search optimization, curriculum design, or something else?

Clarifying these will help me tailor the taxonomy more effectively.

# Response

1.The taxonomy should encode as much generalized information as possible. So yes, all of these are included, but also include others if you think it's relevant.
2. output first as a detailed list of explanations, then also as a simple list, without explanation, but with numbered codes.  e.g. 1. for top level and 1.1, 1.2 ...  for second level codes.
3. it is meant for tagging/classification.
\end{lstlisting}

\subsection{Theme Labeling Prompt}

\begin{lstlisting}[style=plaintext]
# System Prompt

The following series of prompts are for the same problem but at different word limits. For each prompt, summarize what information is included in the prompt by assigning a code from the list below to the information that is included.

Use the list of taxonomies below to guide your analysis. You should only report the information that is included in the prompt using the second-level taxonomies. For example, if the prompt includes "1.1 Task Goal" and "1.2 Scope & Assumptions", you should report "1.1 Task Goal" and "1.2 Scope & Assumptions".

Report your findings in a json file. e.g.:
```json
[
    {
        "prompt": "<fill the prompt here>",
        "word_limit": 10, 
        "taxonomies": [ "1.1 Task Goal", ... ]
    },
    ...
]
```

## List of taxonomies

### 1. Functional Specification

*Describes what the code must do and the exact data it consumes and produces.*

* 1.1 Task Goal - One-sentence statement of the required outcome.
* 1.2 Scope & Assumptions - Preconditions or problem constraints (e.g., "input list is sorted").
* 1.3 Input Specification - Types, structures, and constraints on inputs (e.g., "n <= 10^5, non-negative integers").
* 1.4 Output Specification - Return type and format of the output (e.g., "boolean value, list of strings").
* 1.5 Core Behaviour - Essential functional steps the code must perform, usually mapped to a problem domain.

### 2. Constraints & Robustness

*Sets performance and correctness boundaries for safe, efficient behavior.*

* 2.1 Complexity Targets - Desired time and space complexity (e.g., "<= O(n log n)", "in-place").
* 2.2 Environment Constraints - Platform, language, or hardware requirements (e.g., "no recursion due to stack limit").
* 2.3 Edge Case Handling - Explicit mention of inputs like empty arrays, max values, or special formats.
* 2.4 Error Handling - Required exception behavior, validation, or fallback logic.
* 2.5 Data Invariants - Conditions that must hold true before/after execution (e.g., "list remains sorted").

### 3. Solution Structure & Design Guidance

*Guides how the solution should be implemented or structured.*

* 3.1 Specifying Algorithmic Strategy - Recommends a general technique (e.g., brute-force, recursion, DP, greedy).
* 3.2 Specifying Data Structure - Recommends a structure (e.g., array, set, tree, heap) to enable efficient access.
* 3.3 Forbidden Techniques - Prohibits certain APIs, heuristics, or styles (e.g., "don't use sorting").
* 3.4 Implementation Sketch or Step Breakdown - Provides a sequence of logical steps or pseudocode.
* 3.5 Common Implementation Patterns - Highlights structural motifs (e.g., prefix sum, two-pointer, hash map).
* 3.6 Role or Persona Framing - Adopts a tone or style based on audience (e.g., "explain like I'm a beginner").

### 4. Verification & Integration

*Specifies how correctness is tested and how the code fits into a larger system.*

* 4.1 Sample I/O Pairs - Concrete examples showing expected outputs for given inputs.
* 4.2 Unit Tests or Oracle Checks - Lists or refers to test cases that must pass.
* 4.3 Integration Context - Describes where/how the code will be called or embedded.
* 4.4 Dependencies - External libraries, packages, or imports required (e.g., "uses `collections.Counter`").

# User Prompt

[LLM Summaries at 7 different levels]
\end{lstlisting}

\section{Sample \partialordereval{} Prompts}
\label{sec:app-sample-prompts}

\begin{lstlisting}[style=plaintext]
## Original Prompt


/* forward declare fft. computes fourier
transform in-place */
void
fft(std::vector<std::complex<double>>
&x);

/* Compute the inverse fourier
transform of x in-place.
   Example:
input: [1.0, 1.0, 1.0, 1.0, 0.0, 0.0,
0.0, 0.0]
   output: [{0.5,0},
{0.125,0.301777}, {0,-0},
{0.125,0.0517767}, {0,-0},
{0.125,-0.0517767}, {0,-0},
{0.125,-0.301777}]
*/
void
ifft(std::vector<std::complex<double>>
&x) {




## Original Solution Description


**Step 1: Conjugate All Complex Numbers
in the Input Vector**

The function
begins by applying the complex conjugate
operation to each element in the input
vector of complex numbers. The purpose
of this step is to prepare the data for
the inverse Fourier transform by
leveraging the mathematical property
that the inverse discrete Fourier
transform (IDFT) can be computed using
the forward discrete Fourier transform
(DFT) if the input is conjugated before
and after the transformation. This is
achieved using a standard algorithm from
the C++ Standard Library, which applies
the `std::conj` function to each
element, effectively negating the
imaginary part of every complex number
in the vector. This operation is
performed in-place, ensuring that no
additional memory is allocated and the
original data is directly modified.
Special care is taken to ensure that the
transformation is applied to all
elements, regardless of their initial
values, and that the operation is
efficient for vectors of any size.

---
**Step 2: Perform the Forward Fast
Fourier Transform (FFT) In-Place**
After conjugating the input, the
function proceeds to compute the forward
FFT on the modified vector. The FFT is
an efficient algorithm for computing the
DFT, reducing the computational
complexity from O(N^2) to O(N log N),
where N is the size of the input. By
calling the FFT function in-place, the
function avoids unnecessary copying and
leverages the optimized structure of the
input vector. The FFT function itself is
assumed to handle all the necessary bit-
reversal and butterfly operations
internally, ensuring that the frequency-
domain representation is computed
correctly. This step is crucial because,
due to the initial conjugation, the
result of the forward FFT at this stage
is mathematically equivalent to the
conjugate of the inverse FFT of the
original data.

---

**Step 3: Conjugate
the Complex Numbers Again**

Once the
forward FFT is complete, the function
applies the complex conjugate operation
to each element of the transformed
vector a second time. The goal of this
step is to reverse the initial
conjugation and complete the
mathematical equivalence to the inverse
FFT. By conjugating the data both before
and after the forward FFT, the function
effectively computes the IDFT using only
the forward FFT algorithm. This approach
is particularly useful in environments
where only a forward FFT implementation
is available or when code reuse is
desired. The use of the standard
library's transformation algorithm
ensures that this operation is performed
efficiently and in-place, maintaining
the integrity and performance of the
function.

---

**Step 4: Scale the
Output by the Inverse of the Vector
Size**

The final step involves
normalizing the output by dividing each
element of the vector by the total
number of elements, N. This scaling is
necessary because the forward FFT and
its inverse differ by a normalization
factor; specifically, the inverse FFT
requires division by N to ensure that
the transformation is mathematically
correct and that the output matches the
expected amplitude of the original time-
domain signal. The function accomplishes
this using another in-place
transformation, dividing each complex
number by the size of the vector. This
step is essential for correctness, as
omitting it would result in output
values that are N times larger than
intended. Care is taken to use a
floating-point division to avoid integer
truncation, and the operation is applied
uniformly to all elements, ensuring
consistent scaling across the entire
output.

---

This sequence of steps
allows the function to compute the
inverse Fourier transform efficiently
and accurately, using only the forward
FFT implementation and standard C++
library facilities for in-place data
manipulation.




## LLM Summary at length 10


Use forward FFT with conjugation and
scaling for inverse transform.




## LLM Summary at length 25


Computes inverse FFT using forward FFT:
conjugate input, apply FFT, conjugate
again, scale by 1/N. In-place operations
ensure efficiency.




## LLM Summary at length 50


The ifft function computes the inverse
Fourier transform by conjugating the
input, applying the forward FFT,
conjugating again, and scaling by 1/N.
This method uses the existing FFT
implementation, ensuring in-place
operations and mathematical correctness
through conjugation and normalization
steps.




## LLM Summary at length 75


The inverse FFT function computes the
inverse transform using the forward FFT.
It conjugates the input vector, applies
the FFT, conjugates again, and scales by
1/N. This leverages the mathematical
equivalence between forward FFT and
inverse FFT via conjugation. All
operations are in-place, ensuring
efficiency and avoiding memory
allocation. The scaling ensures correct
amplitude normalization, making the
result equivalent to the standard
inverse FFT. This method efficiently
computes the inverse using only the
forward FFT implementation and standard
C++ operations.




## LLM Summary at length 100


The `ifft` function computes the inverse
FFT using a forward FFT implementation.
It first conjugates the input vector
elements, then applies the forward FFT
in-place. After the FFT, the result is
conjugated again to reverse the initial
step, achieving mathematical equivalence
to the inverse transform. Finally, the
output is scaled by dividing each
element by the vector size `N` to
normalize the amplitude. This approach
leverages the relationship between
forward and inverse FFTs, ensuring
correctness without requiring a separate
inverse FFT implementation. All
operations are performed in-place,
optimizing memory usage and
computational efficiency. The
conjugation and scaling steps ensure the
output matches the expected time-domain
signal, making the function both
accurate and resource-efficient.




## LLM Summary at length 150


The `ifft` function computes the inverse
FFT using the forward FFT by conjugating
the input vector, performing the FFT,
conjugating the result again, and
scaling by 1/N. First, all complex
numbers are conjugated to prepare for
the inverse transform. The forward FFT
is then applied in-place, leveraging its
efficiency. After the FFT, the result is
conjugated again to reverse the initial
conjugation, completing the mathematical
equivalence to the inverse FFT. Finally,
each element is scaled by 1/N to
normalize the output, ensuring correct
amplitude. This approach uses in-place
operations to minimize memory usage and
relies on the forward FFT
implementation, making it efficient and
suitable for environments where only a
forward FFT is available. The steps
ensure mathematical correctness while
maintaining performance through
optimized in-place transformations.




## LLM Summary at length 200


The provided solution computes the
inverse Fourier transform (IFFT) using a
forward FFT implementation, leveraging
mathematical properties to avoid direct
implementation of the inverse transform.
The process begins by conjugating all
complex numbers in the input vector,
preparing the data for the forward FFT.
This conjugation ensures that the
subsequent FFT operation aligns with the
mathematical equivalence between the
inverse DFT and the conjugated forward
DFT. Next, the forward FFT is applied
in-place to the conjugated data,
efficiently transforming the signal into
the frequency domain. After the FFT, the
result is conjugated again to reverse
the initial conjugation, completing the
mathematical transformation equivalent
to the inverse FFT. Finally, the output
is scaled by dividing each element by
the vector's size (N), ensuring correct
amplitude normalization. This step is
critical as the forward FFT inherently
lacks the normalization factor required
for the inverse transform. The entire
process is optimized for in-place
operations, minimizing memory usage and
computational overhead. By reusing the
existing FFT implementation, the
solution achieves efficiency and
correctness while adhering to standard
C++ practices. The approach is
mathematically rigorous, ensuring the
output matches the expected time-domain
signal after inverse transformation.
This method is particularly valuable in
scenarios where only a forward FFT is
available, enabling code reuse and
performance optimization.




## Paragraph Sampling at ratio 0.2


Once the forward FFT is complete, the
function applies the complex conjugate
operation to each element of the
transformed vector a second time. The
goal of this step is to reverse the
initial conjugation and complete the
mathematical equivalence to the inverse
FFT. By conjugating the data both before
and after the forward FFT, the function
effectively computes the IDFT using only
the forward FFT algorithm. This approach
is particularly useful in environments
where only a forward FFT implementation
is available or when code reuse is
desired. The use of the standard
library's transformation algorithm
ensures that this operation is performed
efficiently and in-place, maintaining
the integrity and performance of the
function.

---

**Step 4: Scale the
Output by the Inverse of the Vector
Size**
*/
void
ifft(std::vector<std::complex<double>>
&x)




## Paragraph Sampling at ratio 0.4


Once the forward FFT is complete, the
function applies the complex conjugate
operation to each element of the
transformed vector a second time. The
goal of this step is to reverse the
initial conjugation and complete the
mathematical equivalence to the inverse
FFT. By conjugating the data both before
and after the forward FFT, the function
effectively computes the IDFT using only
the forward FFT algorithm. This approach
is particularly useful in environments
where only a forward FFT implementation
is available or when code reuse is
desired. The use of the standard
library's transformation algorithm
ensures that this operation is performed
efficiently and in-place, maintaining
the integrity and performance of the
function.

---

**Step 4: Scale the
Output by the Inverse of the Vector
Size**

The function begins by applying
the complex conjugate operation to each
element in the input vector of complex
numbers. The purpose of this step is to
prepare the data for the inverse Fourier
transform by leveraging the mathematical
property that the inverse discrete
Fourier transform (IDFT) can be computed
using the forward discrete Fourier
transform (DFT) if the input is
conjugated before and after the
transformation. This is achieved using a
standard algorithm from the C++ Standard
Library, which applies the `std::conj`
function to each element, effectively
negating the imaginary part of every
complex number in the vector. This
operation is performed in-place,
ensuring that no additional memory is
allocated and the original data is
directly modified. Special care is taken
to ensure that the transformation is
applied to all elements, regardless of
their initial values, and that the
operation is efficient for vectors of
any size.

---

**Step 2: Perform the
Forward Fast Fourier Transform (FFT) In-
Place**
*/
void
ifft(std::vector<std::complex<double>>
&x)




## Paragraph Sampling at ratio 0.6


Once the forward FFT is complete, the
function applies the complex conjugate
operation to each element of the
transformed vector a second time. The
goal of this step is to reverse the
initial conjugation and complete the
mathematical equivalence to the inverse
FFT. By conjugating the data both before
and after the forward FFT, the function
effectively computes the IDFT using only
the forward FFT algorithm. This approach
is particularly useful in environments
where only a forward FFT implementation
is available or when code reuse is
desired. The use of the standard
library's transformation algorithm
ensures that this operation is performed
efficiently and in-place, maintaining
the integrity and performance of the
function.

---

**Step 4: Scale the
Output by the Inverse of the Vector
Size**

The function begins by applying
the complex conjugate operation to each
element in the input vector of complex
numbers. The purpose of this step is to
prepare the data for the inverse Fourier
transform by leveraging the mathematical
property that the inverse discrete
Fourier transform (IDFT) can be computed
using the forward discrete Fourier
transform (DFT) if the input is
conjugated before and after the
transformation. This is achieved using a
standard algorithm from the C++ Standard
Library, which applies the `std::conj`
function to each element, effectively
negating the imaginary part of every
complex number in the vector. This
operation is performed in-place,
ensuring that no additional memory is
allocated and the original data is
directly modified. Special care is taken
to ensure that the transformation is
applied to all elements, regardless of
their initial values, and that the
operation is efficient for vectors of
any size.

---

**Step 2: Perform the
Forward Fast Fourier Transform (FFT) In-
Place**

After conjugating the input,
the function proceeds to compute the
forward FFT on the modified vector. The
FFT is an efficient algorithm for
computing the DFT, reducing the
computational complexity from O(N^2) to
O(N log N), where N is the size of the
input. By calling the FFT function in-
place, the function avoids unnecessary
copying and leverages the optimized
structure of the input vector. The FFT
function itself is assumed to handle all
the necessary bit-reversal and butterfly
operations internally, ensuring that the
frequency-domain representation is
computed correctly. This step is crucial
because, due to the initial conjugation,
the result of the forward FFT at this
stage is mathematically equivalent to
the conjugate of the inverse FFT of the
original data.

---

**Step 3: Conjugate
the Complex Numbers Again**
*/
void
ifft(std::vector<std::complex<double>>
&x)




## Paragraph Sampling at ratio 0.8


Once the forward FFT is complete, the
function applies the complex conjugate
operation to each element of the
transformed vector a second time. The
goal of this step is to reverse the
initial conjugation and complete the
mathematical equivalence to the inverse
FFT. By conjugating the data both before
and after the forward FFT, the function
effectively computes the IDFT using only
the forward FFT algorithm. This approach
is particularly useful in environments
where only a forward FFT implementation
is available or when code reuse is
desired. The use of the standard
library's transformation algorithm
ensures that this operation is performed
efficiently and in-place, maintaining
the integrity and performance of the
function.

---

**Step 4: Scale the
Output by the Inverse of the Vector
Size**

The function begins by applying
the complex conjugate operation to each
element in the input vector of complex
numbers. The purpose of this step is to
prepare the data for the inverse Fourier
transform by leveraging the mathematical
property that the inverse discrete
Fourier transform (IDFT) can be computed
using the forward discrete Fourier
transform (DFT) if the input is
conjugated before and after the
transformation. This is achieved using a
standard algorithm from the C++ Standard
Library, which applies the `std::conj`
function to each element, effectively
negating the imaginary part of every
complex number in the vector. This
operation is performed in-place,
ensuring that no additional memory is
allocated and the original data is
directly modified. Special care is taken
to ensure that the transformation is
applied to all elements, regardless of
their initial values, and that the
operation is efficient for vectors of
any size.

---

**Step 2: Perform the
Forward Fast Fourier Transform (FFT) In-
Place**

After conjugating the input,
the function proceeds to compute the
forward FFT on the modified vector. The
FFT is an efficient algorithm for
computing the DFT, reducing the
computational complexity from O(N^2) to
O(N log N), where N is the size of the
input. By calling the FFT function in-
place, the function avoids unnecessary
copying and leverages the optimized
structure of the input vector. The FFT
function itself is assumed to handle all
the necessary bit-reversal and butterfly
operations internally, ensuring that the
frequency-domain representation is
computed correctly. This step is crucial
because, due to the initial conjugation,
the result of the forward FFT at this
stage is mathematically equivalent to
the conjugate of the inverse FFT of the
original data.

---

**Step 3: Conjugate
the Complex Numbers Again**

The final
step involves normalizing the output by
dividing each element of the vector by
the total number of elements, N. This
scaling is necessary because the forward
FFT and its inverse differ by a
normalization factor; specifically, the
inverse FFT requires division by N to
ensure that the transformation is
mathematically correct and that the
output matches the expected amplitude of
the original time-domain signal. The
function accomplishes this using another
in-place transformation, dividing each
complex number by the size of the
vector. This step is essential for
correctness, as omitting it would result
in output values that are N times larger
than intended. Care is taken to use a
floating-point division to avoid integer
truncation, and the operation is applied
uniformly to all elements, ensuring
consistent scaling across the entire
output.

---

This sequence of steps
allows the function to compute the
inverse Fourier transform efficiently
and accurately, using only the forward
FFT implementation and standard C++
library facilities for in-place data
manipulation.
*/
void
ifft(std::vector<std::complex<double>>
&x)




## Sentence Block Masking at ratio 0.2


/*
[REDACTED]. Special care is taken to
ensure that the transformation is
applied to all elements, regardless of
their initial values, and that the
operation is efficient for vectors of
any size.

---

**Step 2: Perform the
Forward Fast Fourier Transform (FFT) In-
Place**

After conjugating the input,
the function proceeds to compute the
forward FFT on the modified vector. The
FFT is an efficient algorithm for
computing the DFT, reducing the
computational complexity from O(N^2) to
O(N log N), where N is the size of the
input. By calling the FFT function in-
place, the function avoids unnecessary
copying and leverages the optimized
structure of the input vector. The FFT
function itself is assumed to handle all
the necessary bit-reversal and butterfly
operations internally, ensuring that the
frequency-domain representation is
computed correctly. This step is crucial
because, due to the initial conjugation,
the result of the forward FFT at this
stage is mathematically equivalent to
the conjugate of the inverse FFT of the
original data.

---

**Step 3: Conjugate
the Complex Numbers Again**

Once the
forward FFT is complete, the function
applies the complex conjugate operation
to each element of the transformed
vector a second time. The goal of this
step is to reverse the initial
conjugation and complete the
mathematical equivalence to the inverse
FFT. By conjugating the data both before
and after the forward FFT, the function
effectively computes the IDFT using only
the forward FFT algorithm. This approach
is particularly useful in environments
where only a forward FFT implementation
is available or when code reuse is
desired. The use of the standard
library's transformation algorithm
ensures that this operation is performed
efficiently and in-place, maintaining
the integrity and performance of the
function.

---

**Step 4: Scale the
Output by the Inverse of the Vector
Size**

The final step involves
normalizing the output by dividing each
element of the vector by the total
number of elements, N. This scaling is
necessary because the forward FFT and
its inverse differ by a normalization
factor; specifically, the inverse FFT
requires division by N to ensure that
the transformation is mathematically
correct and that the output matches the
expected amplitude of the original time-
domain signal. The function accomplishes
this using another in-place
transformation, dividing each complex
number by the size of the vector. This
step is essential for correctness, as
omitting it would result in output
values that are N times larger than
intended. Care is taken to use a
floating-point division to avoid integer
truncation, and the operation is applied
uniformly to all elements, ensuring
consistent scaling across the entire
output.

---

This sequence of steps
allows the function to compute the
inverse Fourier transform efficiently
and accurately, using only the forward
FFT implementation and standard C++
library facilities for in-place data
manipulation.
*/
void
ifft(std::vector<std::complex<double>>
&x) {




## Sentence Block Masking at ratio 0.4


/*
[REDACTED]. The FFT function itself
is assumed to handle all the necessary
bit-reversal and butterfly operations
internally, ensuring that the frequency-
domain representation is computed
correctly. This step is crucial because,
due to the initial conjugation, the
result of the forward FFT at this stage
is mathematically equivalent to the
conjugate of the inverse FFT of the
original data.

---

**Step 3: Conjugate
the Complex Numbers Again**

Once the
forward FFT is complete, the function
applies the complex conjugate operation
to each element of the transformed
vector a second time. The goal of this
step is to reverse the initial
conjugation and complete the
mathematical equivalence to the inverse
FFT. By conjugating the data both before
and after the forward FFT, the function
effectively computes the IDFT using only
the forward FFT algorithm. This approach
is particularly useful in environments
where only a forward FFT implementation
is available or when code reuse is
desired. The use of the standard
library's transformation algorithm
ensures that this operation is performed
efficiently and in-place, maintaining
the integrity and performance of the
function.

---

**Step 4: Scale the
Output by the Inverse of the Vector
Size**

The final step involves
normalizing the output by dividing each
element of the vector by the total
number of elements, N. This scaling is
necessary because the forward FFT and
its inverse differ by a normalization
factor; specifically, the inverse FFT
requires division by N to ensure that
the transformation is mathematically
correct and that the output matches the
expected amplitude of the original time-
domain signal. The function accomplishes
this using another in-place
transformation, dividing each complex
number by the size of the vector. This
step is essential for correctness, as
omitting it would result in output
values that are N times larger than
intended. Care is taken to use a
floating-point division to avoid integer
truncation, and the operation is applied
uniformly to all elements, ensuring
consistent scaling across the entire
output.

---

This sequence of steps
allows the function to compute the
inverse Fourier transform efficiently
and accurately, using only the forward
FFT implementation and standard C++
library facilities for in-place data
manipulation.
*/
void
ifft(std::vector<std::complex<double>>
&x) {




## Sentence Block Masking at ratio 0.6


/*
[REDACTED]. This approach is
particularly useful in environments
where only a forward FFT implementation
is available or when code reuse is
desired. The use of the standard
library's transformation algorithm
ensures that this operation is performed
efficiently and in-place, maintaining
the integrity and performance of the
function.

---

**Step 4: Scale the
Output by the Inverse of the Vector
Size**

The final step involves
normalizing the output by dividing each
element of the vector by the total
number of elements, N. This scaling is
necessary because the forward FFT and
its inverse differ by a normalization
factor; specifically, the inverse FFT
requires division by N to ensure that
the transformation is mathematically
correct and that the output matches the
expected amplitude of the original time-
domain signal. The function accomplishes
this using another in-place
transformation, dividing each complex
number by the size of the vector. This
step is essential for correctness, as
omitting it would result in output
values that are N times larger than
intended. Care is taken to use a
floating-point division to avoid integer
truncation, and the operation is applied
uniformly to all elements, ensuring
consistent scaling across the entire
output.

---

This sequence of steps
allows the function to compute the
inverse Fourier transform efficiently
and accurately, using only the forward
FFT implementation and standard C++
library facilities for in-place data
manipulation.
*/
void
ifft(std::vector<std::complex<double>>
&x) {




## Sentence Block Masking at ratio 0.8


/*
[REDACTED]. The function accomplishes
this using another in-place
transformation, dividing each complex
number by the size of the vector. This
step is essential for correctness, as
omitting it would result in output
values that are N times larger than
intended. Care is taken to use a
floating-point division to avoid integer
truncation, and the operation is applied
uniformly to all elements, ensuring
consistent scaling across the entire
output.

---

This sequence of steps
allows the function to compute the
inverse Fourier transform efficiently
and accurately, using only the forward
FFT implementation and standard C++
library facilities for in-place data
manipulation.
*/
void
ifft(std::vector<std::complex<double>>
&x) {
\end{lstlisting}

\section{Taxonomy of Prompt Details}
\label{sec:taxonomy-of-prompt-details}

\subsection{Functional Specification}

\textit{Describes what the code must do and the exact data it consumes and produces.}

\begin{enumerate}[label=1.\arabic*]
    \item \textbf{Task Goal} – One-sentence statement of the required outcome.
    \item \textbf{Scope \& Assumptions} – Preconditions or problem constraints (e.g., "input list is sorted").
    \item \textbf{Input Specification} – Types, structures, and constraints on inputs (e.g., $n \leq 10^5$ non-negative integers").
    \item \textbf{Output Specification} – Return type and format of the output (e.g., "boolean value, list of strings").
    \item \textbf{Core Behaviour} – Essential functional steps the code must perform, usually mapped to a problem domain.
\end{enumerate}

\subsection{Constraints \& Robustness}

\textit{Sets performance and correctness boundaries for safe, efficient behavior.}

\begin{enumerate}[label=2.\arabic*]
    \item \textbf{Complexity Targets} – Desired time and space complexity (e.g., "$\leq O(n \log n)$", "in-place").
    \item \textbf{Environment Constraints} – Platform, language, or hardware requirements (e.g., "no recursion due to stack limit").
    \item \textbf{Edge Case Handling} – Explicit mention of inputs like empty arrays, max values, or special formats.
    \item \textbf{Error Handling} – Required exception behavior, validation, or fallback logic.
    \item \textbf{Data Invariants} – Conditions that must hold true before/after execution (e.g., "list remains sorted").
\end{enumerate}

\subsection{Solution Structure \& Design Guidance}

\textit{Guides how the solution should be implemented or structured.}

\begin{enumerate}[label=3.\arabic*]
    \item \textbf{Specifying Algorithmic Strategy} – Recommends a general technique (e.g., brute-force, recursion, DP, greedy).
    \item \textbf{Specifying Data Structure} – Recommends a structure (e.g., array, set, tree, heap) to enable efficient access.
    \item \textbf{Forbidden Techniques} – Prohibits certain APIs, heuristics, or styles (e.g., "don't use sorting").
    \item \textbf{Implementation Sketch or Step Breakdown} – Provides a sequence of logical steps or pseudocode.
    \item \textbf{Common Implementation Patterns} – Highlights structural motifs (e.g., prefix sum, two-pointer, hash map).
    \item \textbf{Role or Persona Framing} – Adopts a tone or style based on audience (e.g., "explain like I'm a beginner").
\end{enumerate}

\subsection{Verification \& Integration}

\textit{Specifies how correctness is tested and how the code fits into a larger system.}

\begin{enumerate}[label=4.\arabic*]
    \item \textbf{Sample I/O Pairs} – Concrete examples showing expected outputs for given inputs.
    \item \textbf{Unit Tests or Oracle Checks} – Lists or refers to test cases that must pass.
    \item \textbf{Integration Context} – Describes where/how the code will be called or embedded.
    \item \textbf{Dependencies} – External libraries, packages, or imports required (e.g., "uses \texttt{collections.Counter}").
\end{enumerate}

\begin{figure*}
    \centering
    \includegraphics[width=0.9\textwidth]{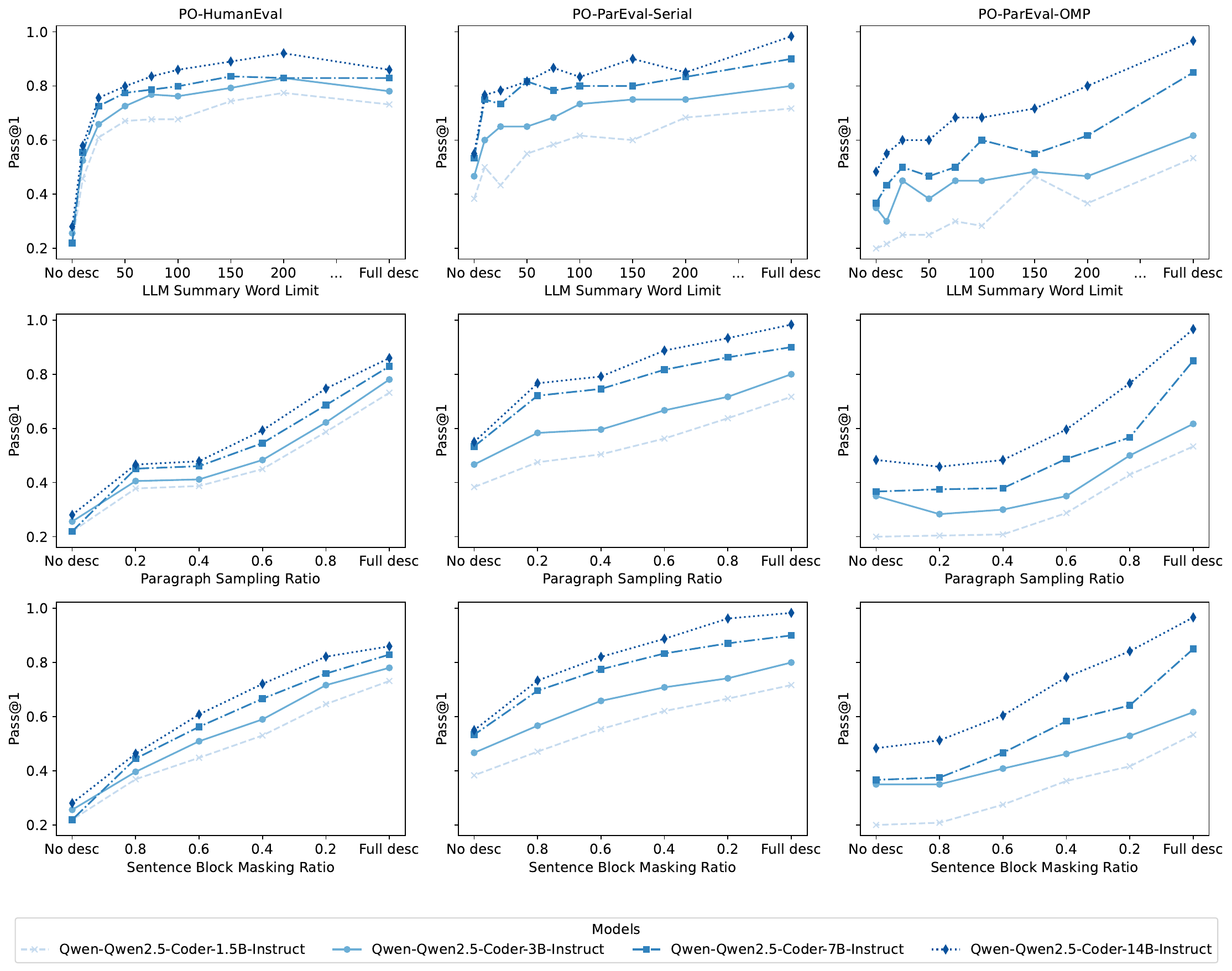}
    \includegraphics[width=0.9\textwidth]{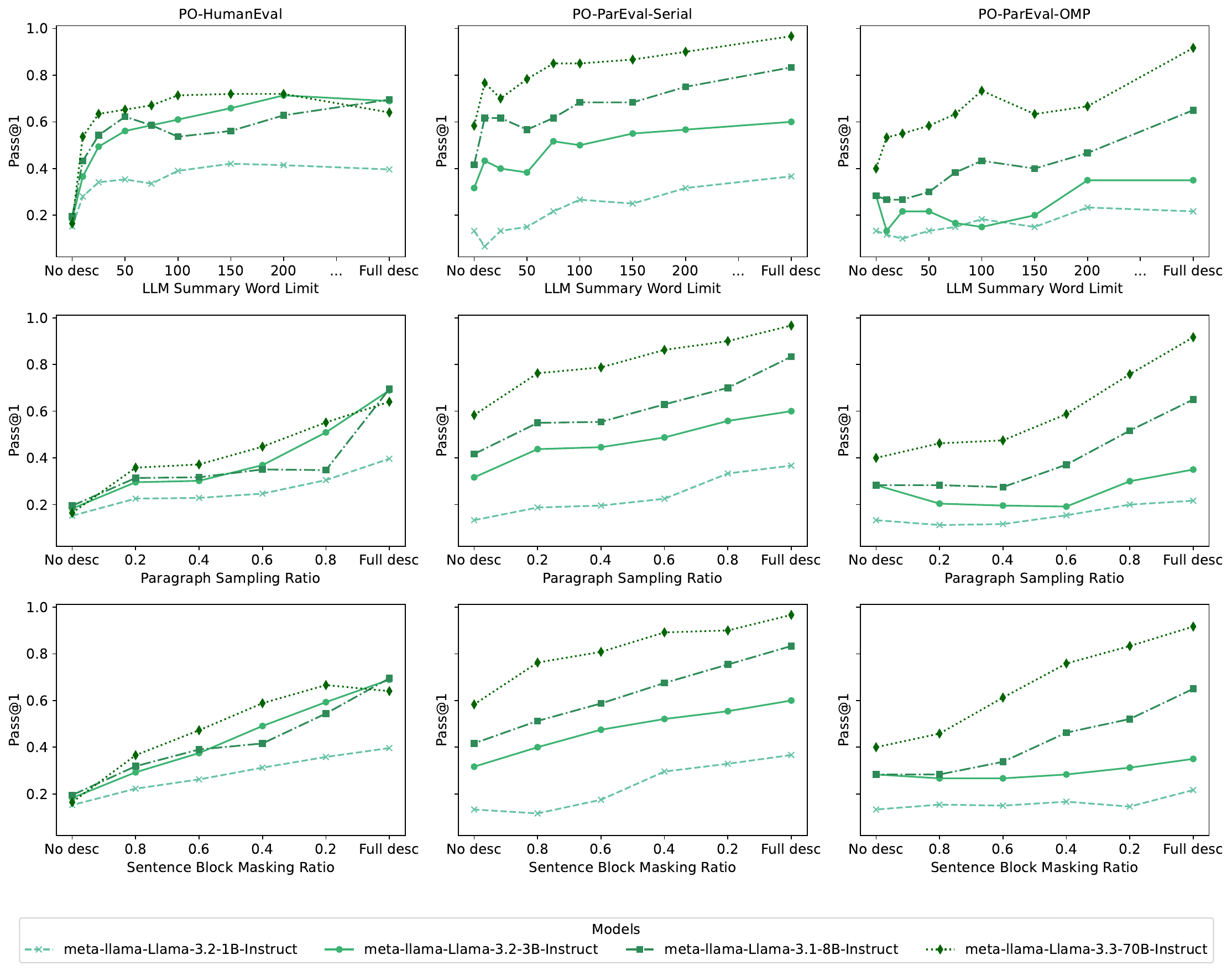}
    \caption{Complete Result of select Qwen2.5-Coder and Meta-Llama3 series models on all source dataset and augmentations. Each row represents one augmentation and each column represents one source dataset. The $x$ axis stand for the prompt detail (with the right hand side means more details), and $y$ axis is pass@1 score for the model at the detail.}
    \label{fig:main-results}
\end{figure*}

\begin{figure*}[t]
  \centering
  % Top‐left
  \begin{subfigure}[b]{0.45\textwidth}
    \includegraphics[width=\linewidth]{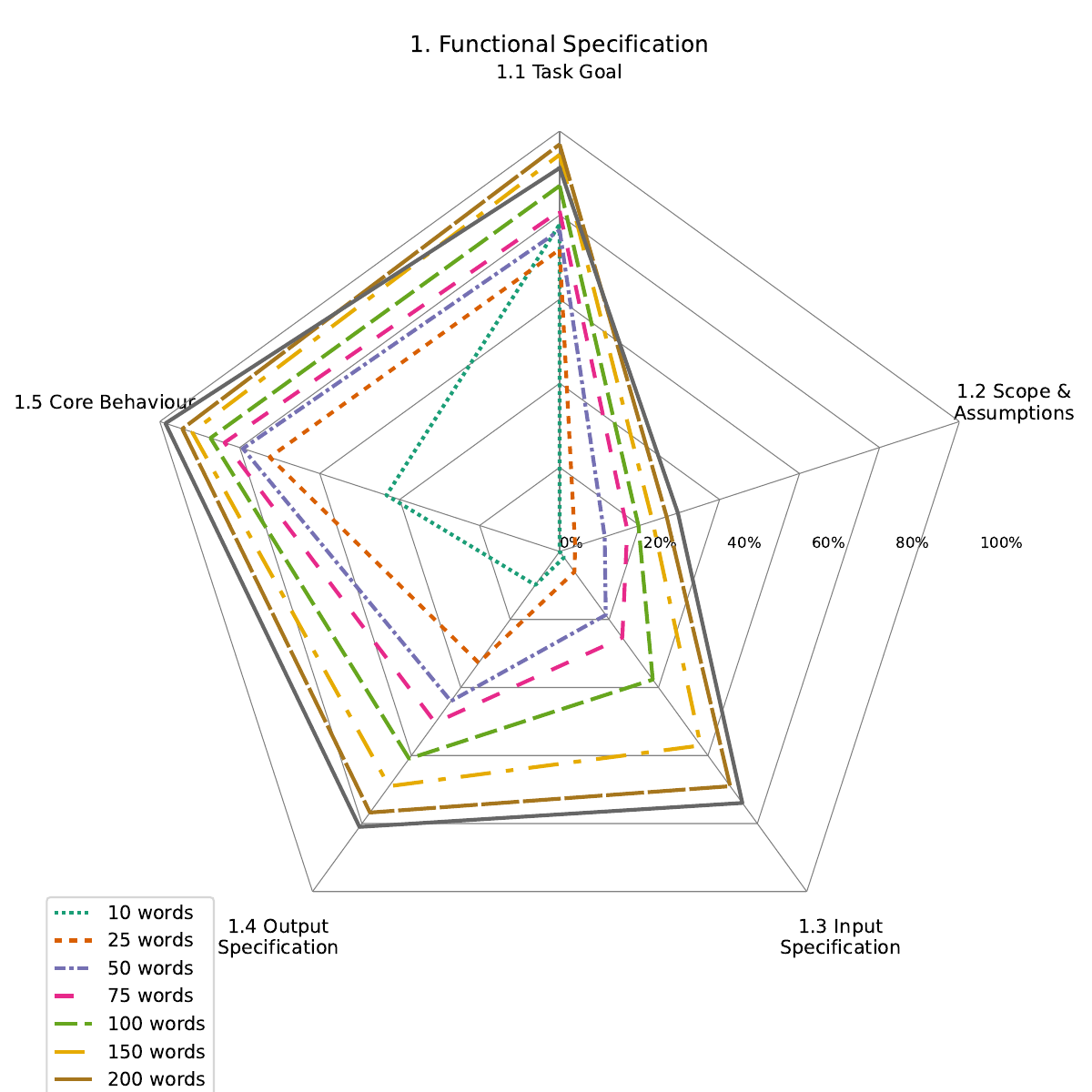}
    %\caption{Caption 1}
    \label{fig:count-of-tags-in-taxonomy-1}
  \end{subfigure}
  \hfill
  % Top‐right
  \begin{subfigure}[b]{0.45\textwidth}
    \includegraphics[width=\linewidth]{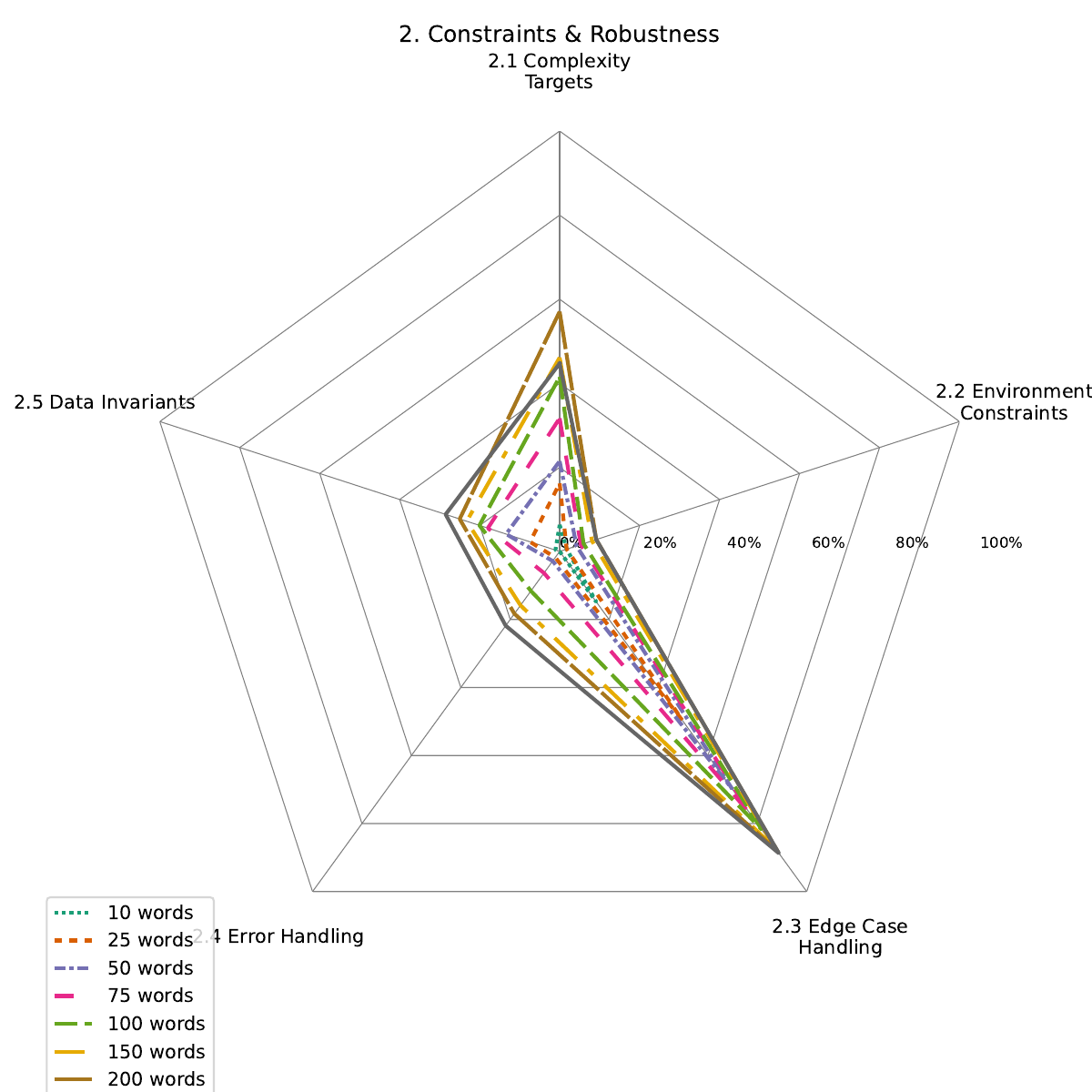}
    %\caption{Caption 2}
    \label{fig:count-of-tags-in-taxonomy-2}
  \end{subfigure}

  % Bottom‐left
  \begin{subfigure}[b]{0.45\textwidth}
    \includegraphics[width=\linewidth]{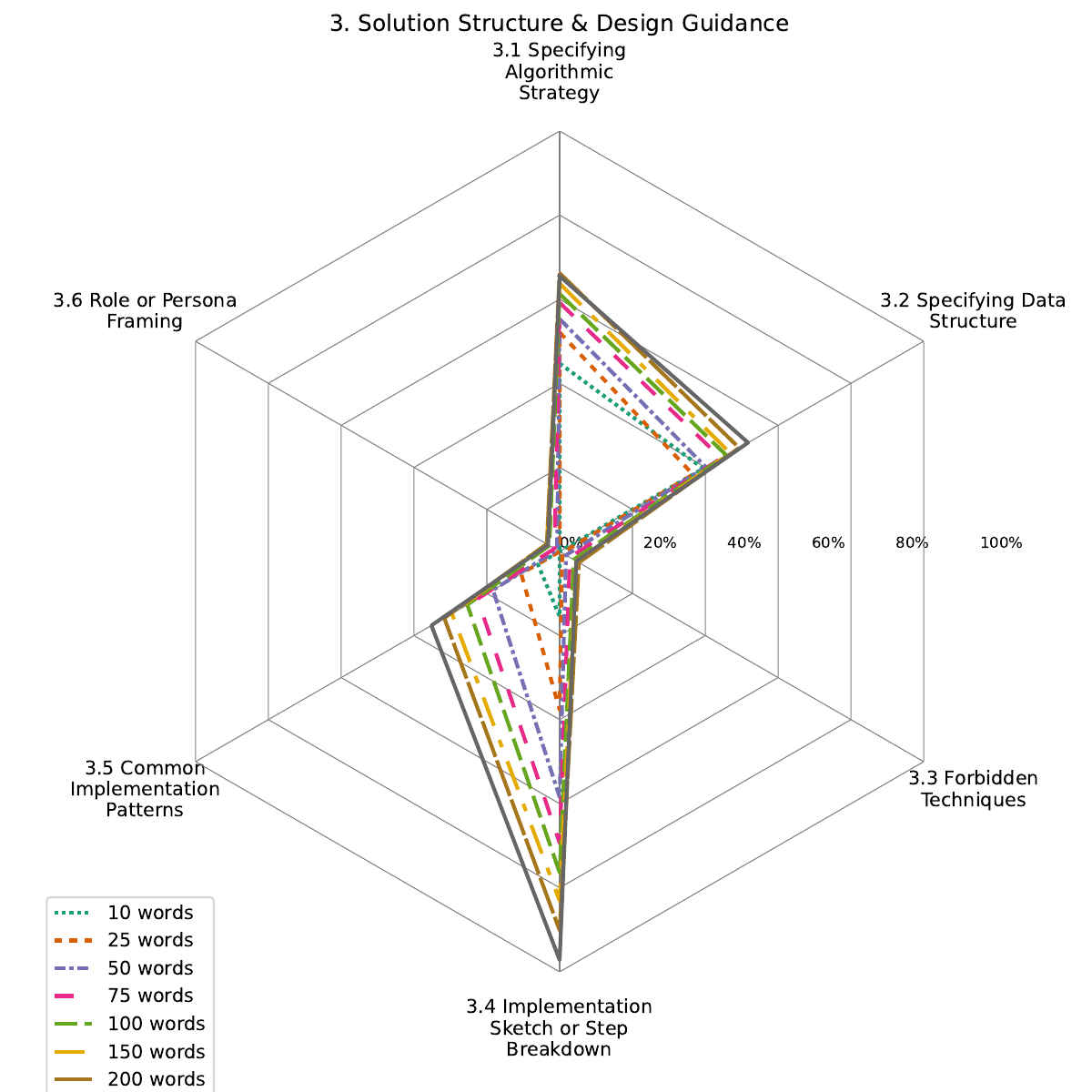}
    %\caption{Caption 3}
    \label{fig:count-of-tags-in-taxonomy-3}
  \end{subfigure}
  \hfill
  % Bottom‐right
  \begin{subfigure}[b]{0.45\textwidth}
    \includegraphics[width=\linewidth]{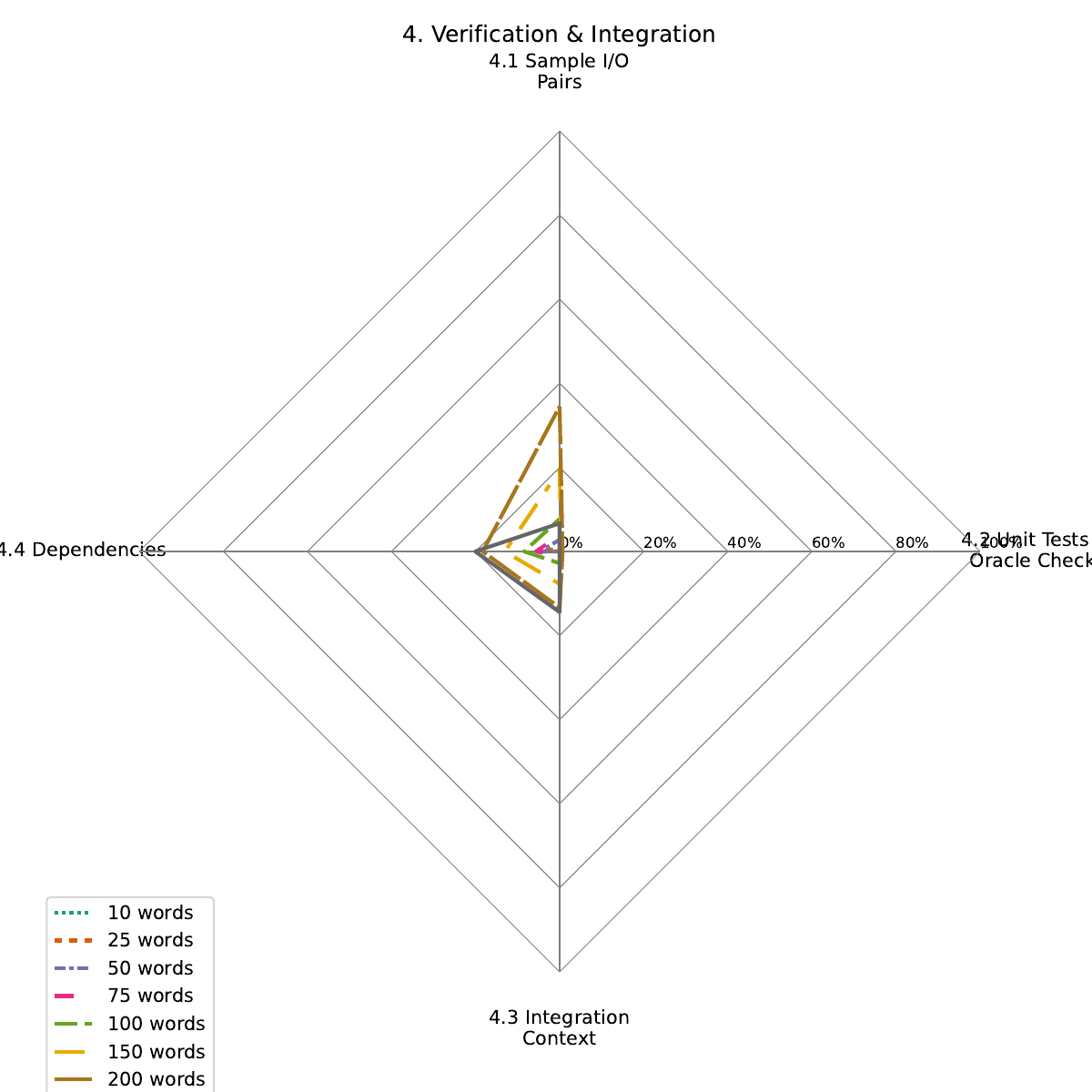}
    %\caption{Caption 4}
    \label{fig:count-of-tags-in-taxonomy-4}
  \end{subfigure}

  \caption{Radar Chats of all LLM Summary taxonomy themes, with one plot per category.}
  \label{fig:count-of-tags-in-taxonomy}
\end{figure*}

\end{document}